\pdfoutput=1

\documentclass[11pt]{article}

\usepackage[final]{acl}
\usepackage{tabularx}
\usepackage[utf8]{inputenc}
\usepackage[arabic,english]{babel}  
\usepackage{textcomp}
\usepackage{longtable}
\usepackage{subcaption}
\usepackage{placeins} 
\usepackage{times}
\usepackage{latexsym}
\usepackage{booktabs}
\usepackage[T1]{fontenc}

\usepackage{multirow}
\renewcommand{\arraystretch}{1.2}
\usepackage{colortbl}
\usepackage{amssymb}

\usepackage{arydshln}
\usepackage{nicematrix}
\usepackage{microtype}
\usepackage{inconsolata}

\usepackage{graphicx}
\usepackage{hyperref}

\title{
\raisebox{-2.1ex}{\protect\includegraphics[height=4.2\fontcharht\font`\B]{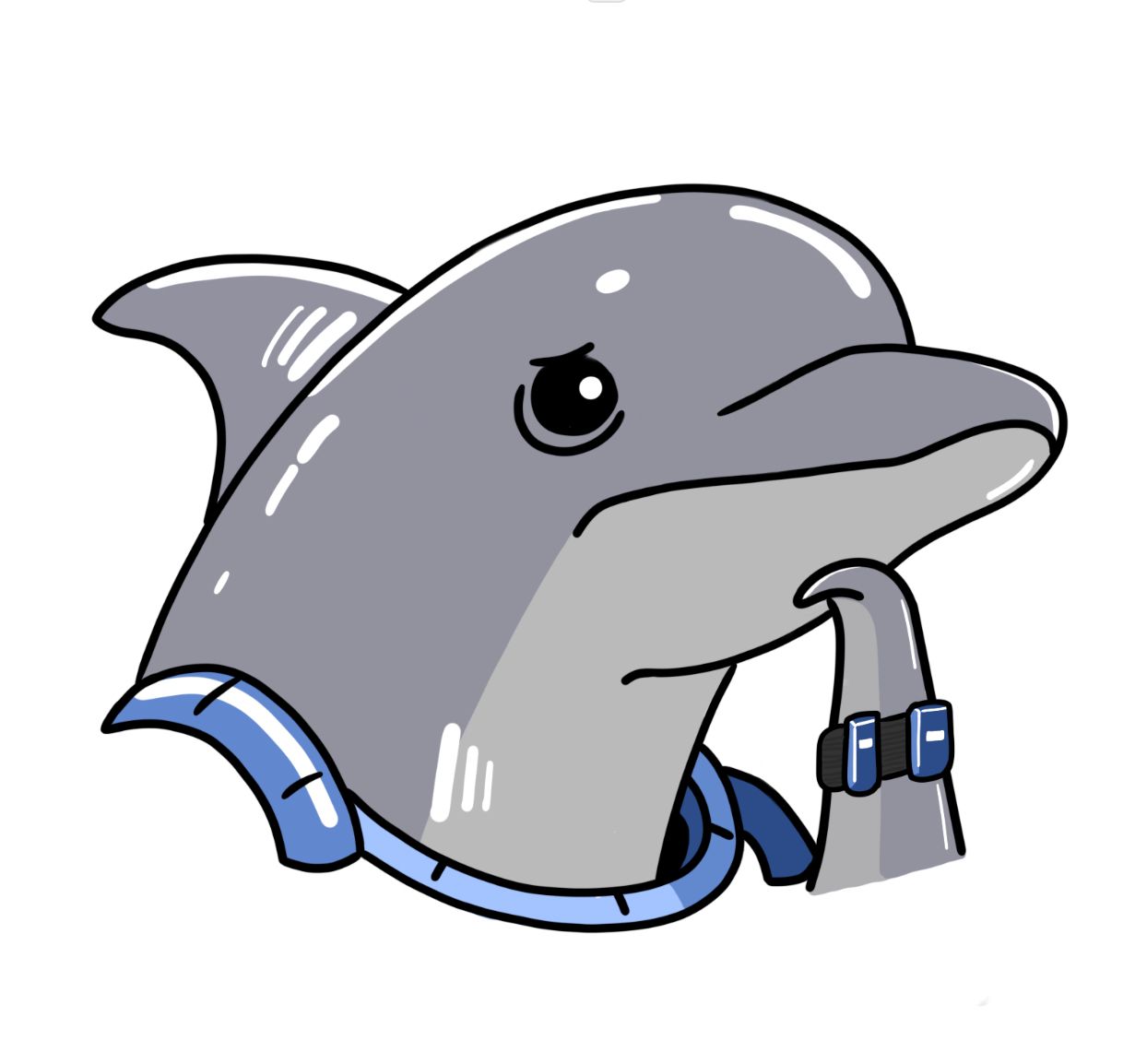}} AraHalluEval: A Fine-grained Hallucination Evaluation Framework for Arabic LLMs}

 \author{Aisha Alansari \and Hamzah Luqman\\
       Information and Computer Science Department, King Fahd University of Petroleum and Minerals \\
       SDAIA-KFUPM Joint Research Center for Artificial Intelligence
       }

\begin{document}

\maketitle


\section*{\centering ABSTRACT}
Recently, extensive research on the hallucination of the large language models (LLMs) has mainly focused on the English language. Despite the growing number of multilingual and Arabic-specific LLMs, evaluating LLMs' hallucination in the Arabic context remains relatively underexplored. The knowledge gap is particularly pressing given Arabic’s widespread use across many regions and its importance in global communication and media. This paper presents the first comprehensive hallucination evaluation of Arabic and multilingual LLMs on two critical Arabic natural language generation tasks: generative question answering (GQA) and summarization. This study evaluates a total of 12 LLMs, including 4 Arabic pre-trained models, 4 multilingual models, and 4 reasoning-based models. To assess the factual consistency and faithfulness of LLMs' outputs, we developed a fine-grained hallucination evaluation framework consisting of 12 fine-grained hallucination indicators that represent the varying characteristics of each task. The results reveal that factual hallucinations are more prevalent than faithfulness errors across all models and tasks. Notably, the Arabic pre-trained model Allam consistently demonstrates lower hallucination rates than multilingual models and a comparative performance with reasoning-based models. The code is available at: \href{https://github.com/aishaalansari57/AraHalluEval}{Github link}.

\section{Introduction}

The emergence of large language models (LLMs) has marked a new era in natural language processing (NLP). LLMs demonstrate exceptional competence in generating coherent and contextually relevant text in multiple languages \cite{chang2024survey}. However, hallucination remains a critical issue for LLMs. Hallucination happens when LLM generates outputs that are factually inaccurate, nonsensical, or misleading \cite{maynez2020faithfulness}. This issue not only undermines their trustworthiness but also limits their practical use in real-world applications.

\begin{figure}[t]
    \centering
    \includegraphics[width=\linewidth]{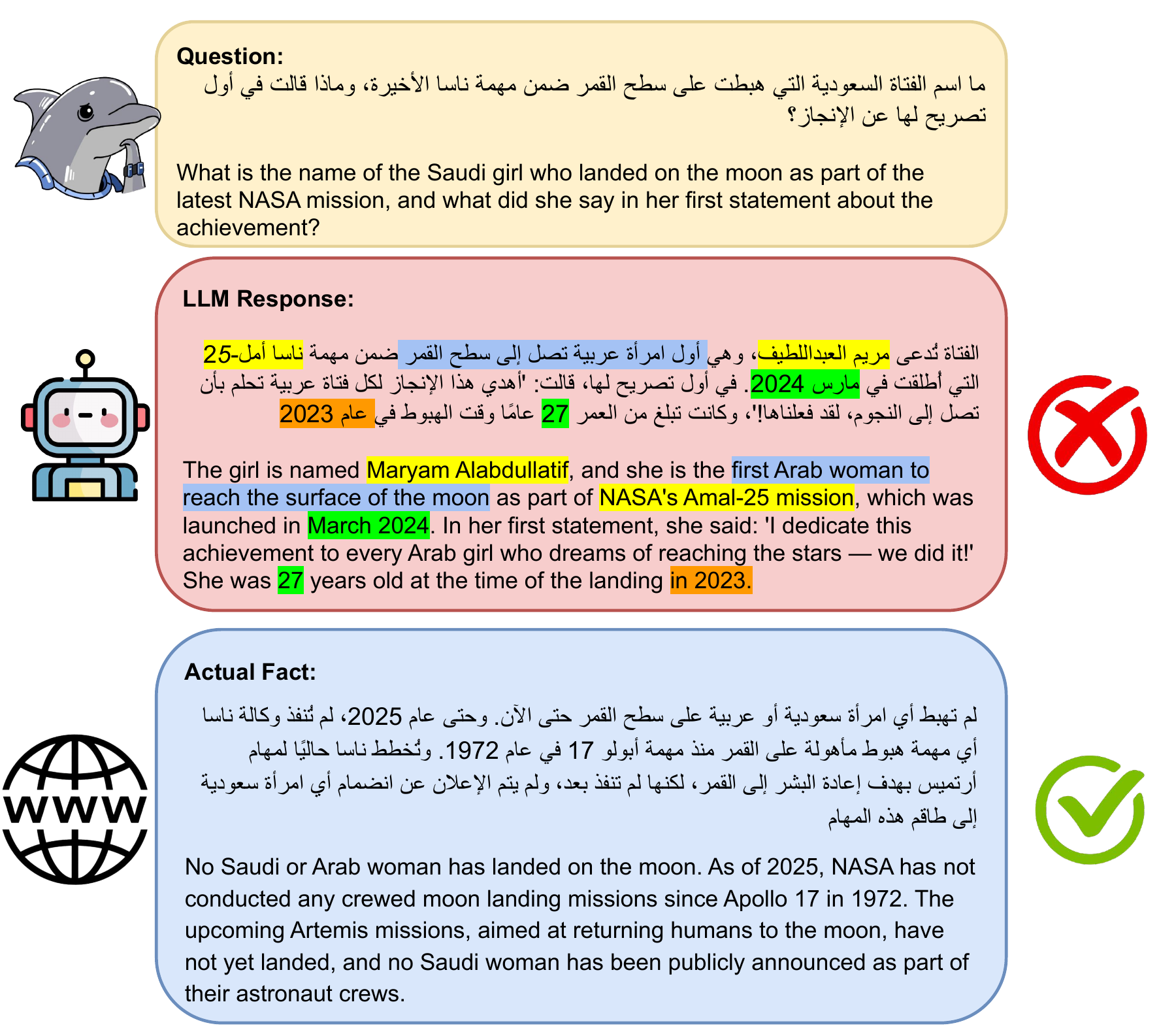}
    \caption{An example of LLM  hallucination errors in the GQA task. \colorbox{yellow}{Named-entity} error denotes incorrect names of people, places, or organizations, \colorbox{green}{value} error denotes wrong dates, ages, or time references, \colorbox{cyan!50}{factual contradiction} represents information not present in the real-world, whereas \colorbox{orange}{response conflict} represents contradicting information within the response itself.}
    \label{fig:new_abstract}
\end{figure}


Hallucination is classified into factual and faithful \cite{huang2025survey}. Factuality hallucination describes the divergence between produced content and known real-world facts, often appearing as factual inconsistency or fabrication. On the other hand, faithfulness hallucination refers to the divergence from the input or context, misaligning with user instructions or internal consistency. 
Figure \ref{abstract} illustrates an example of hallucination in Arabic Generative Question Answering (GQA). In this example, the model introduces named-entity errors (e.g., incorrect names), value errors (e.g., wrong dates), factual contradictions (e.g., claims not supported by real-world facts), and response conflicts (e.g., internal contradictions within the generated response). 

Extensive research on hallucinations in LLMs has predominantly focused on high-resource languages, such as English and Chinese \cite{chang2024survey,huang2025survey}. 
Evaluating LLMs' hallucination in the Arabic context remains relatively underexplored despite the growing number of multilingual and Arabic-specific LLMs \cite{bari2024allam, sengupta2023jais}. Arabic presents unique linguistic challenges due to its morphological richness, complex syntax, and diversity of dialects \cite{farghaly2009arabic, habash2010introduction}. These challenges make hallucination evaluation more complex and necessitate specialized benchmarks and \textcolor{black}{\cite{mubarak-etal-2024-halwasa, abdaljalil2025halluverse25finegrainedmultilingualbenchmark}}methodologies \textcolor{black}{\cite{sibaee-etal-2024-asos}}.

To address this limitation, we conduct a comprehensive evaluation of state-of-the-art (SOTA) Arabic and multilingual LLMs on two critical generative tasks: GQA and text summarization. Twelve LLMs have been evaluated in this work. We also evaluated the performance of \textcolor{black}{four} reasoning-based models on the TruthfulQA hallucination benchmark.  Our evaluation goes beyond conventional metrics by incorporating fine-grained human evaluation to assess hallucinations using a multi-dimensional criterion encompassing both factuality and faithfulness. Twelve fine-grained hallucination types have been identified in this study and used to evaluate LLMs. Through this comparative analysis, we identify strengths and shortcomings of the evaluated LLMs in generating factual outputs. 
The main contributions of this study can be summarized as follows:
\begin{itemize}
\item{Propose a multi-dimensional assessment criterion for LLMs' hallucination in Arabic.}
\item{Evaluate hallucination in Arabic, multi-lingual, \textcolor{black}{and reasoning-based} LLMs on Arabic GQA and text Summarization tasks.}
\item{Present a manually annotated dataset for evaluating hallucinations in Arabic LLM outputs across GQA and summarization tasks.}
\item{Compare \textcolor{black}{four} reasoning-based LLMs on the TruthfulQA hallucination benchmark using parallel English and Arabic questions.}

\end{itemize}
\section{Related work}\label{sec:literature}
\paragraph{Hallucination in LLMs. }
Hallucination in LLMs compromises model reliability and poses safety concerns in real-world applications such as healthcare, education, and law. Previous studies have extensively explored hallucination in LLMs within English contexts, focusing primarily on detection and mitigation strategies \cite{ji2023survey,chang2024survey,huang2025survey,rawte2023survey}. 
To mitigate hallucination in LLMs, prior studies proposed strategies, such as self-verification approaches \cite{manakul2023selfcheckgpt}, grounding model outputs in external outputs \cite{lewis2020retrieval}, introducing self-consistency decoding \cite{wang2022self}, and contrastive decoding \cite{chuang2023dola}. 

Despite the advancement in LLMs, hallucination remains understudied in low-resource languages like Arabic. While reasoning-focused models such as GPT-4o \cite{openai2024gpt4ocard} and DeepSeek-R1 \cite{guo2025deepseek} show promise in mitigating hallucinations in English, their effectiveness in Arabic generative tasks is largely unknown. Meanwhile, Arabic-specific LLMs like Jais \cite{sengupta2023jais}, Fanar \cite{team2025fanar}, and Allam \cite{bari2024allam} have been developed, but their hallucination behavior has yet to be systematically evaluated. Given Arabic’s morphological complexity and dialectal variation, dedicated benchmarks are essential for evaluating factuality and faithfulness in Arabic LLM outputs \cite{mubarak-etal-2024-halwasa}. Besides, cross-lingual comparisons between Arabic-focused and multilingual LLMs—such as Gemma3 \cite{team2024gemma}, LLaMA3 \cite{grattafiori2024llama}, and Qwen2.5 \cite{hui2024qwen2}—are crucial for understanding how language-specific features affect hallucination. This evaluation is crucial, as language-specific behaviors may lead to significant differences in hallucination tendencies and factual reliability when generating Arabic content.


\paragraph{Hallucination Evaluation. }
Evaluating hallucination in LLMs is essential to understand their factual reliability and ensure alignment with user intent. Accordingly, another area of research concentrates on assessing the hallucination of models across various NLP tasks. For instance, \citet{maynez2020faithfulness} provided a comprehensive study on hallucinations for abstractive summarization, revealing that SOTA models frequently generate factually and faithfully inconsistent summaries. Their study shows that even summaries with high ROUGE scores can be unfaithful, which highlights the need for better evaluation methods. 

A variety of measures have been developed to evaluate the faithfulness of abstractive summarization.
 The metrics encompass entailment-based measures \cite{kryscinski2020evaluating,goyal2020evaluating, laban2022summac}, as well as question-generation and question-answering metrics \cite{fabbri2022qafacteval,manakul2023mqag,subbiah2024storysumm}. Recently, attention has transitioned to LLM-based metrics \cite{gao2023human,chan2023chateval,song2024finesure} that utilize LLMs to evaluate the fidelity of a summary. 
To evaluate hallucination in GQA, prior research has explored multiple approaches, including fine-tuning LLMs to detect factual inconsistencies \cite{kadavath2022language} and analyzing internal model states to identify hallucinated or factually incorrect claims \cite{farquhar2024detecting,su2024unsupervised}. 

In parallel, several benchmark datasets have been introduced to facilitate standardized evaluation, including TruthfulQA \cite{lin2022truthfulqa}, which targets common misconceptions; FreshQA \cite{vu2024freshllms}, which focuses on time-sensitive knowledge; HaluEval \cite{li2023halueval}, designed for hallucination categorization. These datasets enable a more comprehensive analysis of hallucination tendencies in GQA. Despite these advancements, hallucination evaluation remains largely unexplored in the Arabic language. Most existing benchmarks and evaluation metrics have been developed for English, leaving a significant gap in assessing the factuality and faithfulness of Arabic generative outputs.

Our work bridges this research gap by providing an extensive comparative evaluation of hallucination phenomena in both Arabic-specific, multilingual, and reasoning LLMs on Arabic GQA and summarization tasks. We aim to systematically measure hallucination in LLMs, identify linguistic features contributing to hallucinations, and benchmark reasoning-enhanced models in an Arabic linguistic context.





\section{AraHalluEval Framework}
\label{sec_method}

\begin{figure*}[tbph]
    \centering
    \includegraphics[width=2\columnwidth]{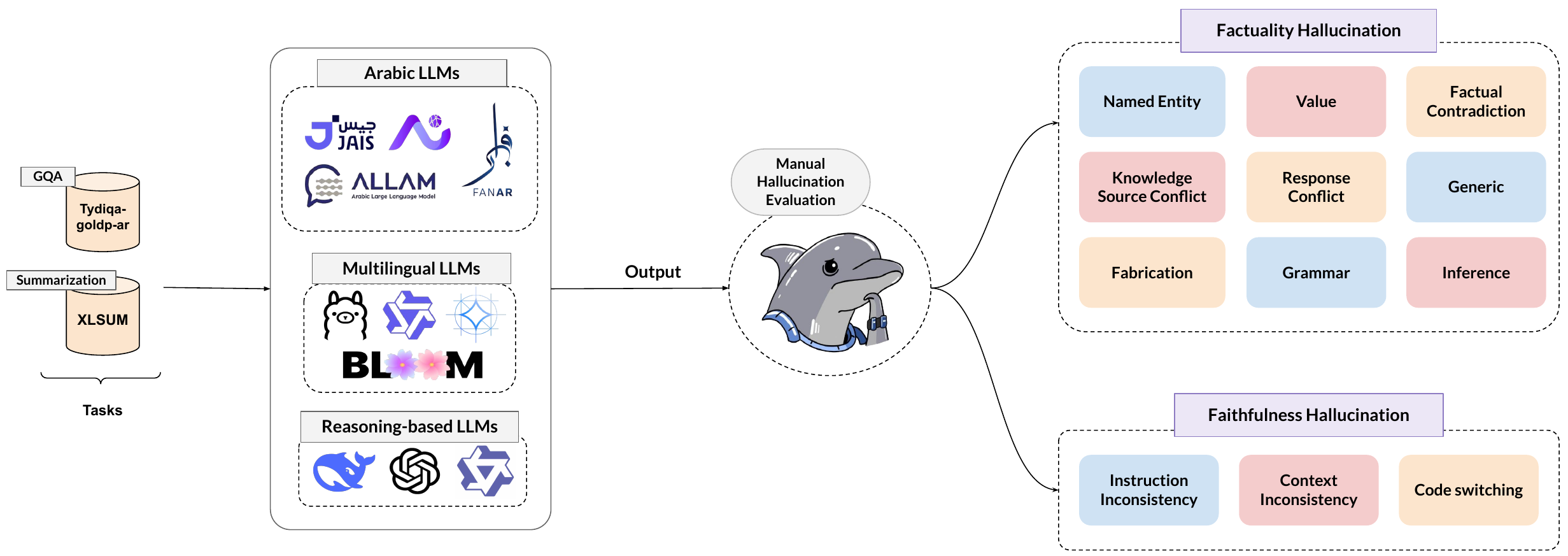}
    \caption{The AraHalluEval pipeline.}
    \label{fig:pipeline}
\end{figure*}

We evaluate the hallucination of Arabic and multilingual LLMs in a zero-shot setup on two tasks: GQA and text summarization. Figure \ref{fig:pipeline} illustrates the hallucination evaluation pipeline.  For each task, we fed the input data to the evaluated LLMs, and their responses were manually evaluated to determine the level of hallucination.

\subsection{Tasks and Datasets}
\paragraph{GQA.} 
This task involves generating natural language answers to open-ended questions. The evaluated models are required to generate accurate, coherent, and contextually faithful answers. For this task, we used the Tydiqa-goldp-ar dataset \cite{clark2020tydi}. The TyDiQA-GoldP-AR dataset is a realistic and challenging benchmark. It aims to replicate genuine human curiosity by having annotators generate questions with minimal background knowledge of the article.
We sampled 300 random questions from the test set of this dataset and fed them into the selected LLMs. Then, the output of each LLM is manually evaluated using nine hallucination indicators to measure its hallucination. We selected this number of samples because using the complete test set is challenging due to its large size and the high cost of human evaluation. 
Moreover, we used the TruthfulQA \cite{lin2022truthfulqa} dataset to evaluate the reasoning-based models. This dataset contains only samples in the English language; therefore, we manually translated them into Arabic to enable cross-lingual comparison. 
More details about the dataset translation process are present in Appendix \ref{truthful}. 

\textbf{Summarization.} This task requires models to generate a concise and faithful summary of longer texts. We randomly sampled 100 instances from the Arabic test set of the XLSum benchmark \cite{hasan2021xl}. This dataset contains high-quality summaries written by professional journalists, making it a suitable benchmark for testing the faithfulness of the LLMs in abstractive summarization. However, this dataset is more challenging for manual annotation compared to GQA, given the number of evaluated LLMs and the long summary of each sample, which justifies the selection of 100 samples from this dataset. 

\subsection{Models Selection}
This study aims to include a wide range of Arabic and multilingual LLMs to evaluate their factuality and faithfulness to Arabic GQA and summarization. Therefore, a total of 12 models were evaluated, of which 4 are Arabic pre-trained LLMs, 4 are multilingual LLMs, and 4 are reasoning-based LLMs.

\paragraph{Arabic LLMs.} In this study, we evaluated the hallucination of the following Arabic LLMs: (1) \textit{Allam-preview-7b-instruct} \cite{bari2024allam}, is an Arabic LLM pre-trained using 4 trillion English tokens followed 1.2 trillion Arabic/English tokens; (2) \textit{Fanar-1-9b} \cite{team2025fanar} developed by pre-training the google/gemma-2-9b model on 1 trillion Arabic and English tokens;  (3) \textit{Jais-6.7b} \cite{sengupta2023jais}, which is a bilingual Arabic-English LLM, optimized for proficiency in Arabic while demonstrating robust capabilities in English;
  (4) \textit{Noon-7b} \cite{naseej_noon2023}, which is an Arabic LLM based on BLOOM, trained using various Arabic tasks. 

\paragraph{Multilingual LLMs.} The hallucination of LLMs that support the Arabic language is also evaluated in this work. We selected the following multilingual LLMs: (1) \textit{LLama3-8b} \cite{grattafiori2024llama}, which is Meta's 8B multilingual model, part of the LLaMA 3 series, trained on a diverse corpus covering over 20 languages, including Arabic; (2) \textit{Qwen2.5-7b} \cite{hui2024qwen2} is the latest series of Qwen large language models which supports 29 languages, including Arabic; (3) \textit{Gemma3-8b} \cite{team2024gemma} is a language model released by Google DeepMind in 2024; and (4) \textit{bloom-7.1b} \cite{le2023bloom} is a multilingual model from BigScience, with 7.1B parameters, trained on 46 languages, including Arabic. 


\paragraph{Reasoning-based LLMs.} We also evaluated reasoning-based models to explore whether models with explicit reasoning capabilities have lower hallucination rates in Arabic tasks. We used the following models: \textcolor{black}{(1) \textit{GPT-4o} \cite{openai2024gpt4ocard}, which is  OpenAI’s native multimodal (“omni”) that generates text, images, and audio for real-time interaction; (2) \textit{GPT-o3} \cite{openai2025o3o4mini} is one of the strongest OpenAI's reasoning models;} (3) \textit{DeepSeek-R1} \cite{guo2025deepseek}, which uses a cold-start supervised fine-tuning for more stable reasoning; (4) \textit{QwQ-32B} \cite{qwq32b}, which is a reasoning model from the Qwen series with 32B parameters. This model demonstrated strong reasoning capabilities using reinforcement learning techniques.

\subsection{Hallucination Evaluation}
To assess the factual consistency and faithfulness of LLMs' outputs, we developed a fine-grained hallucination evaluation framework for the GQA and summarization tasks.  This framework introduces 12 fine-grained hallucination indicators that represent the varying characteristics of each task, as shown in Figure \ref{fig:example}. We used these types to manually evaluate the output of each LLM involved in this study. 
We conducted a manual evaluation of hallucinations using native Arabic speakers, since existing automatic metrics (e.g., ROUGE, BLEU) are insufficient for factual consistency \cite{maynez2020faithfulness}.

\subsubsection{Hallucination Indicators}
\label{sec_hall_indicators}

Each task is evaluated along two core dimensions: factuality and faithfulness. \textit{Factuality hallucination} refers to the discrepancy between generated content and established real-world facts, frequently manifesting as factual inconsistency or fabrication \cite{huang2025survey}. On the other hand, \textit{faithfulness hallucination} refers to the deviation from the user instructions or context, resulting in misalignment with user instructions or internal consistency \cite{huang2025survey}. 

\begin{figure*}[!h]
    \centering
    \includegraphics[width=2.2\columnwidth]{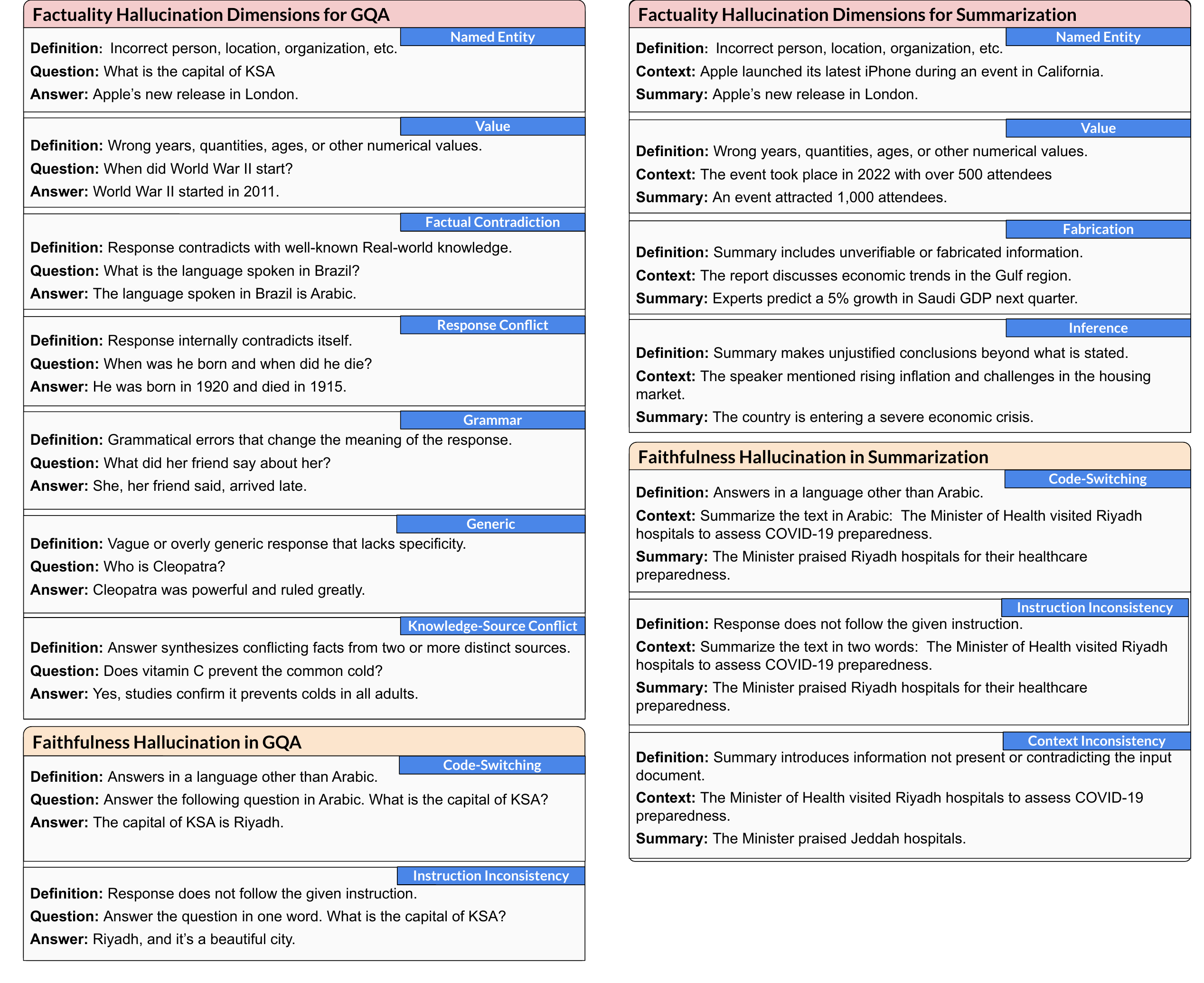}
    \caption{Definitions and examples of the hallucination indicators used to measure the hallucination of each LLM. }
    \label{fig:example}
\end{figure*}

\textbf{GQA}: Hallucination in GQA reflects the model’s failure to produce a factually correct or relevant answer. Therefore, we assess hallucination with respect to real-world knowledge and commonsense plausibility. 
\textit{Factuality} is measured by seven factors: \textit{named-entity}, \textit{value}, \textit{factual contradiction}, \textit{knowledge-source conflict}, \textit{response-conflict},  \textit{generic}, and \textit{grammar}. 
\textit{Faithfulness} is measured by two indicators: \textit{instruction inconsistency}, where the model deviates from the given prompt; and \textit{code-switching}, where the model produces output in a language other than Arabic despite explicit instruction to respond in Arabic. Figure \ref{fig:example} defines and gives an example of each indicator.



\textbf{Summarization}: Hallucination occurs in abstractive summarization when the generated summary contradicts the original text or contains information not available in the source document.
Abstractive summarization's \textit{Factuality} is measured using five indictaor: \textit{named-entity}, \textit{value}, \textit{grammar}, \textit{fabrication}, and \textit{inference}. Models frequently vary in verbosity; some provide longer responses with more details, hence increasing the probability of factual inaccuracies, especially regarding numerical or named-entity references. Therefore,  we present \textit{hallucination density} to ensure that the evaluation is fair for different summary lengths and details provided. It is calculated as the proportion of correct and incorrect facts in each summary. By normalizing hallucination counts relative to the total number of factual units, hallucination density provides a fairer basis for comparing models regardless of how concise or verbose their outputs are. 
The \textit{faithfulness} of the generated text is measured by \textit{instruction inconsistency}, which captures cases where the model fails to follow the input prompt or summary guidelines accurately, \textit{context inconsistency}, which reflect cases where the model's summary contradicts or deviates from the original source content, and \textit{code-switching}, which flags any word written in a language other than Arabic. We also used a \textit{human rating} indicator where annotators rate each summary on a 5-point Likert scale based on how accurately it reflects the original text. Figure \ref{fig:example} defines and gives an example of each indicator.

\subsubsection{Human Evaluation}
Our evaluation process covers 5,600 outputs generated by the evaluated LLMs (300 for GQA and 100 for text summarization tasks generated by 12 LLMs). Given the complexity of hallucination and the lack of reliable automatic metrics, especially for Arabic, we conduct detailed manual annotation for both tasks. Annotations are performed by native Arabic speakers with linguistic and NLP training. Annotators were provided with definitions, examples, and task-specific guidelines for each hallucination type and score.

Each sample is annotated by two independent annotators. Disagreements are resolved by a third expert based on guideline consistency. For GQA, LLM outputs are evaluated for factual correctness based on commonly accepted knowledge (i.e., no context was given to the model). For summarization, the generated summary is evaluated against the original article.  More details about the annotation are present in Appendices \ref{sec:appendix1}, \ref{sec:appendix2}, and \ref{sec:appendix3}.
\section{Results and Discussion}

\begin{table*}[tbph]
\centering
\small
\caption{Hallucination scores on the Arabic GQA task. 
\textcolor{black}{NE = Named-entity errors, Val = Value errors, Contr. = Factual contradictions, Conflic. = Conflict hallucinations, Gramm. = Grammar errors, Gen. = Generic/Imprecise hallucinations, KSC = Knowledge source conflict, Instr. = Instruction inconsistency, CSw. = Code-switching.}}
\label{tab:gqa_hallucination_summary}
\renewcommand{\arraystretch}{1.5}
\setlength{\tabcolsep}{4pt}
\resizebox{\textwidth}{!}{%
\begin{tabular}{lccc|ccccccc|c|cc|c|c}
\hline
\multirow{2}{*}{\textbf{Model}} & \multicolumn{3}{c|}{\textbf{Model Lang.}} & 
\multicolumn{8}{c|}{\textbf{Factual Errors}} 
& \multicolumn{3}{c|}{\textbf{Faithfulness Errors}} 
& \multirow{2}{*}{\textbf{Average}} \\
& \textbf{Arabic} & \textbf{Multi.} & \textbf{Rsn.} & \textbf{NE} & \textbf{Val} & \textbf{Contr.} & \textbf{Conflic.} & \textbf{Gramm.} & \textbf{Gen.} & \textbf{KSC} & \textbf{Total} & \textbf{Instr.} & \textbf{CSw.} & \textbf{Total} & \\
\hline
Allam        & \checkmark &            &            & 0.083 & 0.240 & 0.307 & 0.000 & 0.003 & 0.070 & 0.023 & \textbf{0.727} & 0.007 & 0.030 & \textbf{0.037} & \textbf{0.382} \\
Fanar        & \checkmark &            &            & 0.120 & 0.227 & 0.313 & 0.000 & 0.003 & 0.143 & 0.030 & 0.837 & 0.033 & 0.147 & 0.180 & 0.508 \\
Jais-6.7b    & \checkmark &            &            & 0.137 & 0.103 & 0.240 & 0.000 & 0.000 & 0.527 & 0.003 & 1.010 & 0.480 & 0.063 & 0.543 & 0.777 \\
Noon         & \checkmark &            &            & 0.197 & 0.393 & 0.547 & 0.003 & 0.003 & 0.243 & 0.020 & 1.407 & 0.050 & 0.070 & 0.120 & 0.763 \\ 
\hline
Gemma        &            & \checkmark &            & 0.193 & 0.297 & 0.453 & 0.003 & 0.000 & 0.193 & 0.020 & 1.160 & 0.040 & 0.090 & 0.130 & 0.645 \\
Bloom-7b     &            & \checkmark &            & 0.213 & 0.303 & 0.510 & 0.003 & 0.003 & 0.287 & 0.020 & 1.339 & 0.037 & 0.083 & \textbf{0.120} & 0.730 \\
llama        &            & \checkmark &            & 0.163 & 0.207 & 0.313 & 0.000 & 0.000 & 0.257 & 0.023 & \textbf{0.963} & 0.030 & 0.090 & \textbf{0.120} & \textbf{0.542} \\
qwen2.5-7b   &            & \checkmark &            & 0.220 & 0.267 & 0.300 & 0.003 & 0.003 & 0.310 & 0.030 & 1.133 & 0.060 & 0.117 & 0.177 & 0.655 \\
\hline
DeepSeek-r1  &            & \checkmark & \checkmark & 0.070 & 0.127 & 0.200 & 0.000 & 0.003 & 0.193 & 0.010 & 0.603 & 0.067 & 0.083 & 0.150 & 0.377 \\
GPT-4o       &            & \checkmark & \checkmark & 0.040 & 0.067 & 0.120 & 0.000 & 0.000 & 0.127 & 0.010 & \textbf{0.364} & 0.033 & 0.073 & 0.106 & \textbf{0.235} \\
GPT-o3       &            & \checkmark & \checkmark & 0.050 & 0.083 & 0.130 & 0.000 & 0.003 & 0.137 & 0.010 & 0.413 & 0.030 & 0.067 & \textbf{0.097} & 0.255 \\
QwQ          &            & \checkmark & \checkmark & 0.110 & 0.150 & 0.280 & 0.003 & 0.003 & 0.223 & 0.013 & 0.779 & 0.070 & 0.093 & 0.163 & 0.471 \\
\hline
\end{tabular}
}
\end{table*}

Several experiments have been conducted to evaluate the hallucination of the selected models on Arabic GQA and summarization tasks. More information about the experiment setup and prompts selection is available in Appendix \ref{app:experiments}. 

\paragraph{Models Hallucination. }
Tables \ref{tab:gqa_hallucination_summary} and \ref{tab:summ_hallucination_summary} show the results of evaluated LLMs on Arabic GQA and text summarization tasks, respectively. \textcolor{black}{The average hallucination score is computed as the mean of the total factual and faithfulness hallucinations for each model.} 

Both tables show a clear contrast in performance across Arabic and multilingual LLMs.  As shown in Table \ref{tab:gqa_hallucination_summary}, the best-performing model, Allam, achieved the lowest average hallucination score of 0.382, with minimal faithfulness error rate and factuality. The low factual and faithfulness hallucination error rates of Allam indicate strong adherence to real-world knowledge and user instructions.
 In contrast, models like Noon, Jais, and Bloom exhibit significantly higher hallucination scores, with average scores of 0.777, 0.763, and 0.730, respectively. The high error rates of these models are driven primarily by factual contradictions, named-entity, value, and generic errors, consistent with the general trend that value and named-entity hallucinations dominate in GQA outputs. These errors can be attributed to the models' difficulty in handling time-sensitive or fact-specific questions, compounded by the absence of grounding in real-world temporal knowledge. Faithfulness errors, including instruction inconsistency and code-switching, are relatively rare across models, with Jais being a notable exception, which indicates that this bilingual model may face challenges in maintaining language consistency and adhering to instructions.

Table \ref{tab:summ_hallucination_summary} shows the hallucination error rates of the evaluated models on the text summarization task. For this task, we used ten indicators to measure the hallucination of each LLM. More details about these indicators are available in Section \ref{sec_hall_indicators}.
 As shown in the table, hallucination patterns diverge significantly, where fabrication and context inconsistency being the most prevalent error types across all models. This highlights the models’ tendency to introduce fabricated content or deviate from the original document’s context, which is a major issue in summarization, where it is important for the resulting summary to be close to the source.

Similar to the GQA task, Allam obtained the lowest average hallucination score of 0.215 and achieved the best human rating of 5. These results confirm that its outputs are both factual and faithful. In contrast, Fanar and Gemma exhibit high average hallucination scores of 1.215 and 1.000 for factual hallucinations, respectively. Bloom-7b also received the lowest rate by human evaluators, which indicates a big discrepancy between its output and the context of the original text, which could be attributed to the presence of noisy or low-quality data in Bloom’s pretraining corpus.

\begin{figure}[h!]
    \begin{subfigure}{\textwidth}
\includegraphics[width=0.5\textwidth]{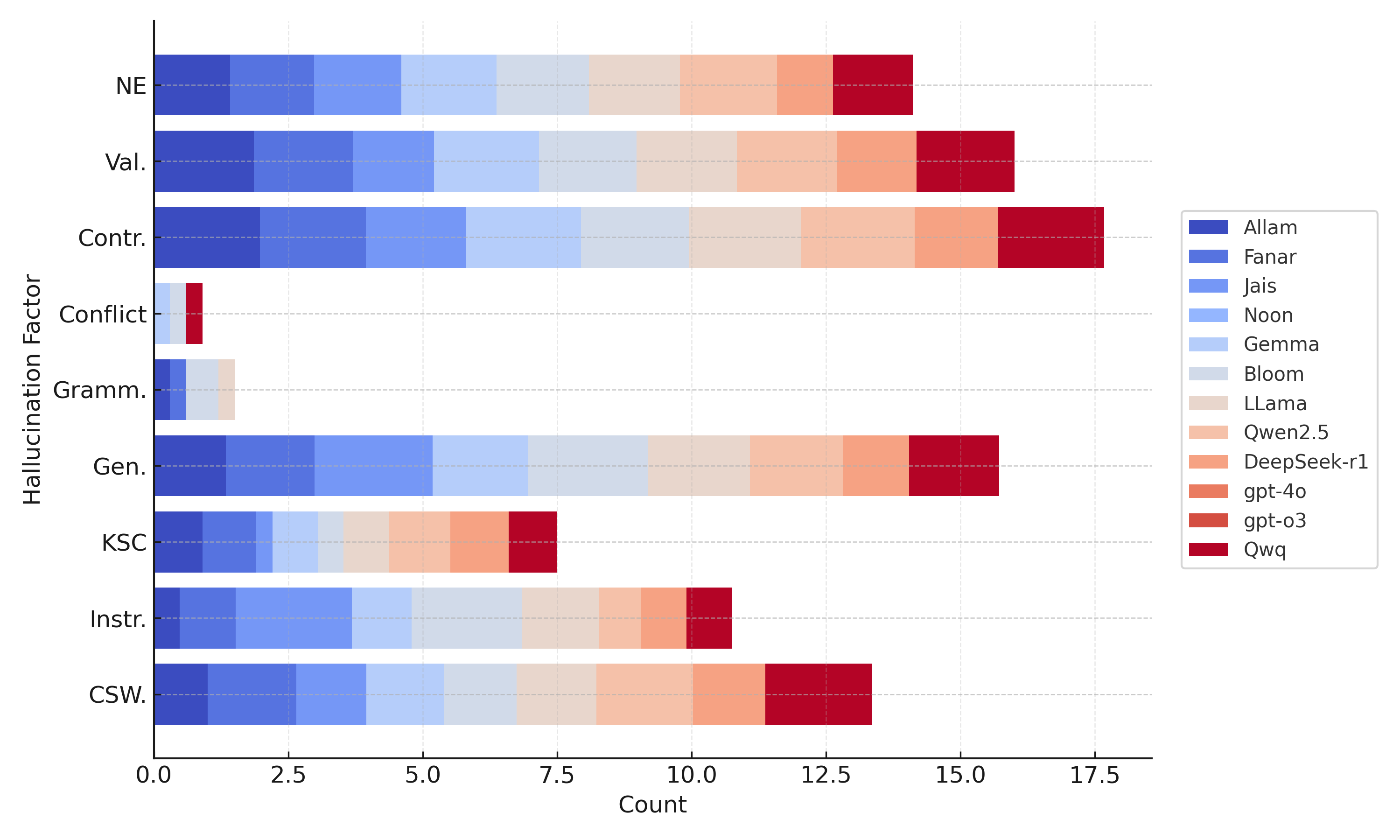} 
        \caption*{(a)}
    \end{subfigure}
      \begin{subfigure}{\textwidth}
\includegraphics[width=0.5\textwidth]{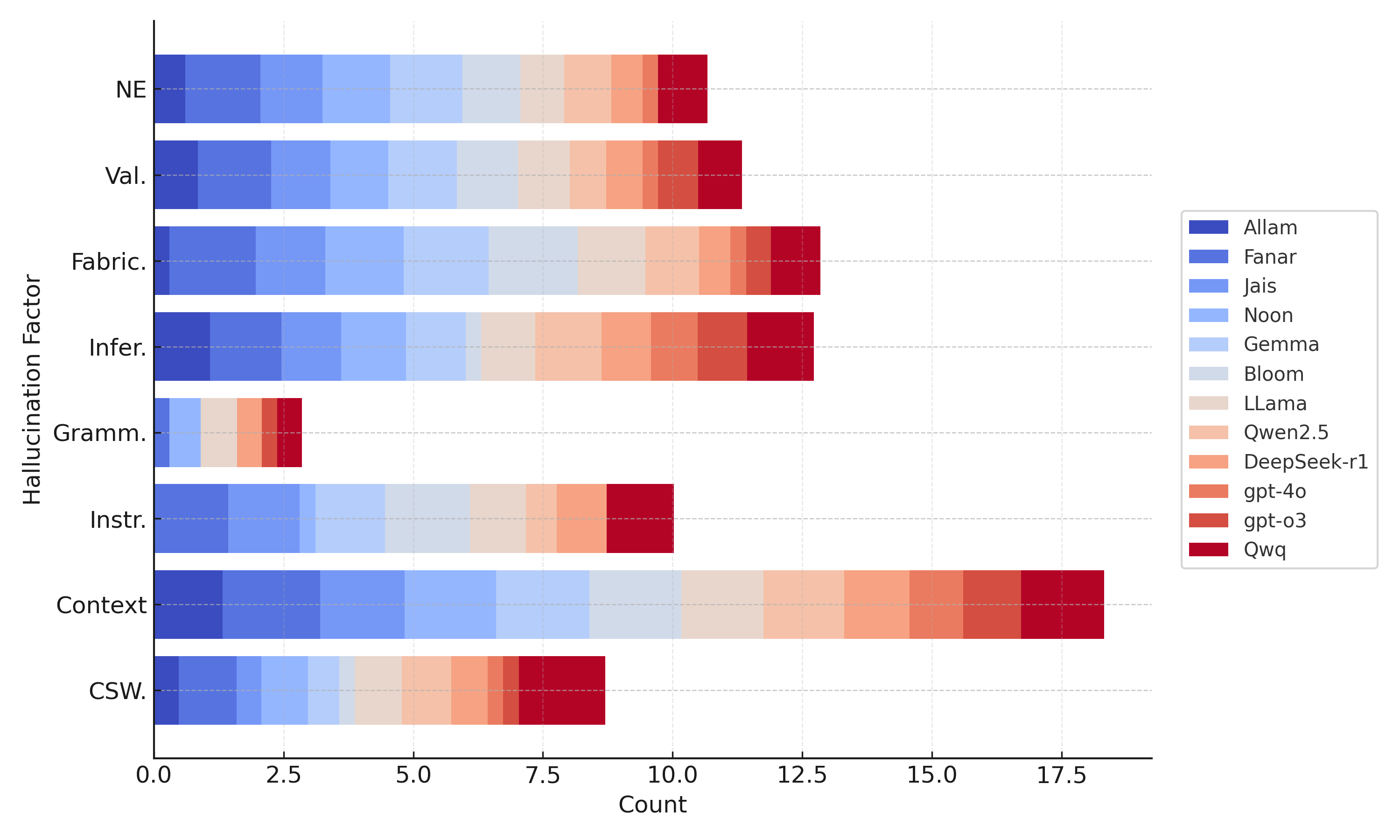} 
        \caption*{(b)}

    \end{subfigure}

    \caption{Frequency of hallucination types (log$_{10}$-scaled) generated by evaluated LLMs across (a) GQA and (b) text summarization tasks.}
    \label{fig:hallucination_factors_comparison}
\end{figure}

\paragraph{Hallucination Indicators. }Figure~\ref{fig:hallucination_factors_comparison} presents the distribution of hallucination types of each LLM in the Arabic GQA and text summarization tasks. In the GQA task (Figure~\ref{fig:hallucination_factors_comparison}a), factual contradiction hallucinations are the most frequent, followed by generic, value, and named-entity hallucinations. These factual errors are the most dominant among the other factors, which show challenges in answering time-sensitive and entity-centric questions. Faithfulness errors, such as instruction inconsistency and code-switching, are also observed but to a lesser extent.

In contrast, the summarization task, as shown in (Figure~\ref{fig:hallucination_factors_comparison}b), shows a different pattern. Context inconsistency and fabrication are the most frequent hallucination types generated by LLMs. This highlights summarization’s susceptibility to content invention and divergence from context. Errors such as inference, value, and named-entity remain common but are less dominant. These differences emphasize how hallucination types vary across NLG tasks and reinforce the need for task-specific evaluation criteria.

\begin{table*}
\centering
\small
\caption{Hallucination scores on the Arabic summarization task. \textcolor{black}{NE = Named-entity errors, Val = Value errors, Fabric. = Fabrications, Infer. = Inference errors, Gramm. = Grammar errors, Instr. = Instruction inconsistency, and CSw. = Code-switching.}}

\label{tab:summ_hallucination_summary}
\renewcommand{\arraystretch}{1.5}
\setlength{\tabcolsep}{4pt}
\resizebox{\textwidth}{!}{%
\begin{tabular}{lccc|ccccc|c|c|ccc|c|c|c}
\hline
\multirow{2}{*}{\textbf{Model}} & \multicolumn{3}{c|}{\textbf{Model Lang.}} & 
\multicolumn{7}{c|}{\textbf{Factual Errors}} 
& \multicolumn{4}{c|}{\textbf{Faithfulness Errors}} 
& \multirow{2}{*}{\textbf{Average}} & \multirow{2}{*}{\textbf{Human Rating}} \\
& \textbf{Arabic} & \textbf{Multi.} & \textbf{Rsn.} & \textbf{NE} & \textbf{Val} & \textbf{Fabric.} & \textbf{Infer.} & \textbf{Gramm.} 
& \textbf{Total} & \textbf{Density} & \textbf{Instr.} & \textbf{Context} & \textbf{CSw.} & \textbf{Total} & & \\
\hline
Allam        & \checkmark &            &            & 0.030 & 0.060 & 0.010 & 0.110 & 0.000 & 0.210 & \textbf{0.066} & 0.000 & 0.200 & 0.020 & 0.220 & \textbf{0.215} & 5 \\
Fanar        & \checkmark &            &            & 0.270 & 0.250 & 0.455 & 0.230 & 0.010 & 1.215 & 0.486 & 0.260 & 0.750 & 0.120 & 1.130 & 1.172 & 3 \\
Jais         & \checkmark &            &            & 0.150 & 0.130 & 0.210 & 0.130 & 0.000 & 0.620 & 0.344 & 0.230 & 0.420 & 0.010 & 0.660 & 0.638 & 3 \\
Noon         & \checkmark &            &            & 0.192 & 0.121 & 0.313 & 0.172 & 0.030 & 0.828 & 0.277 & 0.010 & 0.576 & 0.071 & 0.675 & 0.743 & 4 \\ 
\hline
Gemma        &            & \checkmark &            & 0.240 & 0.200 & 0.430 & 0.130 & 0.000 & 1.000 & 0.410 & 0.210 & 0.610 & 0.030 & 0.850 & 0.925 & 3 \\
Bloom-7b     &            & \checkmark &            & 0.120 & 0.140 & 0.510 & 0.010 & 0.000 & 0.780 & 0.545 & 0.420 & 0.590 & 0.010 & 1.020 & 0.783 & 1 \\
Llama        &            & \checkmark &            & 0.060 & 0.090 & 0.190 & 0.100 & 0.040 & 0.480 & 0.212 & 0.110 & 0.370 & 0.070 & 0.550 & \textbf{0.515} & 3 \\
Qwen2.5      &            & \checkmark &            & 0.070 & 0.040 & 0.100 & 0.180 & 0.000 & 0.390 & \textbf{0.128} & 0.110 & 0.370 & 0.083 & 0.563 & 0.477 & 4 \\
\hline
DeepSeek-r1  &            & \checkmark & \checkmark & 0.030 & 0.040 & 0.030 & 0.080 & 0.020 & 0.200 & 0.075 & 0.080 & 0.170 & 0.040 & 0.290 & 0.245 & 5 \\
GPT-4o       &            & \checkmark & \checkmark & 0.010 & 0.010 & 0.010 & 0.070 & 0.000 & 0.100 & \textbf{0.021} & 0.000 & 0.100 & 0.010 & 0.110 & \textbf{0.105} & 5 \\
GPT-o3       &            & \checkmark & \checkmark & 0.000 & 0.050 & 0.020 & 0.080 & 0.010 & 0.160 & 0.032 & 0.000 & 0.120 & 0.010 & 0.130 & 0.145 & 5 \\
QwQ          &            & \checkmark & \checkmark & 0.080 & 0.060 & 0.080 & 0.180 & 0.020 & 0.420 & 0.147 & 0.190 & 0.390 & 0.460 & 1.040 & 0.730 & 4 \\
\hline
\end{tabular}
}
\end{table*}

\begin{figure}[h]
\includegraphics[width=0.5\textwidth]{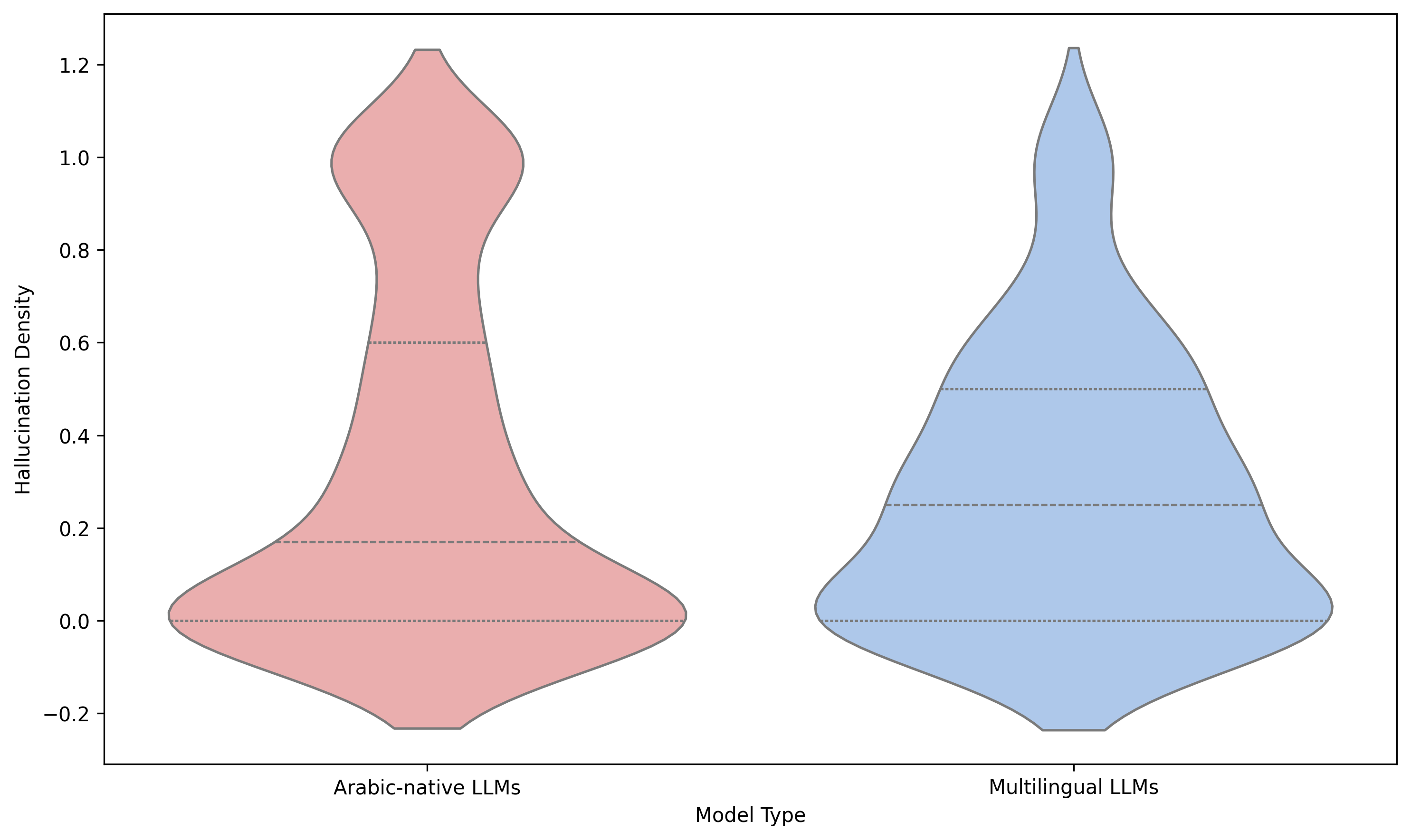} 
  
    \caption{Distribution of hallucination density across Arabic and multilingual LLMs using the summarization task. }
    \label{fig:armul}
\end{figure}

\paragraph{Arabic vs. Multilingual LLMs. }
\textcolor{black}{Figure \ref{fig:armul} shows the hallucination density distribution of the evaluated Arabic and multilingual LLMs on the text summarization task. While the difference in hallucination density between Arabic and multilingual models did not reach statistical significance (t = -1.41, p = 0.161), the trend indicates that Arabic models may produce fewer hallucinations on average. This can be attributed to the small size of the dataset and the number of evaluated LLMs. The results of the paired t-test revealed a statistically significant difference at the 5\% level (p = 0.0186), indicating that Allam produces significantly fewer hallucinations than Qwen2.5-7b. The negative t-statistic further supports this finding, showing that Allam consistently generates summaries with lower hallucination density. This confirms the superior factual faithfulness of Allam in Arabic summarization.}

For GQA, we conducted a Mann-Whitney U test to compare factual hallucination rates between models. When comparing all Arabic models against all multilingual models, the difference was also statistically significant, with a U-statistic of 649,023.5 and a p-value of 8.19e-6 (p < 0.01). These findings indicate that Arabic models are generally more robust in reducing factual hallucinations in the Arabic GQA task compared to their multilingual counterparts. More details about selecting the significance test are present in Appendix \ref{sec:appendix6}.

\begin{table}[ht]
\centering
\caption{Hallucination rates of the reasoning-based models on Arabic and English outputs using the TruthfulQA dataset.}
\label{tab:tqa_hallucination}
\resizebox{.8\columnwidth}{!}{%
\begin{tabular}{llc}
\hline
\textbf{Language} & \textbf{Model} & \textbf{Hallucination Rate} \\
\hline
Arabic  & Allam         & 0.666 \\
        & DeepSeek R1   & 0.519 \\
        & GPT-4o        & 0.448 \\
        & GPT-o3        & 0.649 \\
        & QwQ           & 0.524 \\
        
\hline
English & Allam         & 0.616 \\
        & DeepSeek R1   & 0.482 \\
        & GPT-4o        & 0.425 \\
        & GPT-o3        & 0.548 \\
        & QwQ           & 0.497 \\
        
\hline
\multicolumn{2}{l}{\textbf{t-statistic}} & 3.37 \\
\multicolumn{2}{l}{\textbf{p-value}} & 0.028 \\
\hline
\end{tabular}
}
\end{table}

\paragraph{Reasoning-based models.}
\textcolor{black}{Tables \ref{tab:gqa_hallucination_summary} and \ref{tab:summ_hallucination_summary} show the performance of four reasoning-based models in Arabic GQA and summarization tasks, respectively. As shown in Table \ref{tab:gqa_hallucination_summary}, gpt-4o demonstrate the best factuality and faithfulness scores of 0.364 and 0.235, respectively, whereas QwQ exhibits the highest factual and faithfulness errors of 0.779 and 0.471, respectively. Notably, the Arabic-pretrained model, Allam, rivals reasoning-based models, achieving an average hallucination score of 0.382 with competitive performance to QwQ and DeepSeek-r1, which underscores the effectiveness of language-specific pretraining in mitigating hallucinations.}

\textcolor{black}{A similar trend is shown in table \ref{tab:summ_hallucination_summary}, where gpt-4o attains the best average hallucination score of 0.105, followed by gpt-o3, whereas QwQ exhibits the highest average hallucination score of 0.730. The Arabic pre-trained model, Allam, outperforms DeepSeek-r1 and QwQ with a factual density of 0.066 and a faithfulness score of 0.220, which also underscores the effectiveness of language-specific pretraining.}

\textcolor{black}{Table \ref{tab:tqa_hallucination} shows the hallucination rates of four reasoning-based LLMs: DeepSeek R1, GPT-4o, GPT-o3, and QwQ and the best-performing Arabic-centric model, Allam, when responses are generated in Arabic and English using the TruthfulQA dataset. We used the coarse-grained definition of the hallucination introduced in this dataset, where the generated responses are compared against the ground-truth. Responses that do not match the ground-truth are considered hallucinations. Using this definition, we computed the hallucination rate reported in Table \ref{tab:tqa_hallucination}.  As shown, the hallucination rate is consistently higher in Arabic outputs relative to English outputs across all reasoning-based models. For instance, the GPT-o3 model demonstrates a hallucination rate of 0.649 in Arabic compared to 0.548 in English. Likewise, DeepSeek-r1 and QwQ exhibit higher hallucination rates in Arabic with 0.519 and 0.524, respectively, compared to 0.482 and 0.497 in English. A  two-tailed paired samples t-test indicates a statistically significant difference, with a t-statistic of 3.37 and a p-value of $2.81 \times 10^{-2}$. These findings suggest that reasoning-based LLMs are more prone to generating hallucinations when responding in Arabic, which underscores the need for further study and targeted enhancements in Arabic.}

\section{Conclusion}

In this study, we presented the first comprehensive evaluation of hallucination in Arabic across Arabic and multilingual LLMs using two NLG tasks: GQA and summarization. We proposed a multi-dimensional hallucination evaluation framework that incorporates both factuality and faithfulness, tailored specifically to the challenges of Arabic GQA and summarization. Furthermore, we evaluated the performance of reasoning-based LLMs using the TruthfulQA benchmark with parallel Arabic and English questions and gold answers. Our findings reveal that factual hallucinations are more prevalent than faithfulness errors across all models and tasks. Arabic models consistently produced fewer hallucinations compared to their multilingual counterparts. 
\textcolor{black}{Future work will focus on expanding the evaluation to include additional open-source models and a broader range of NLG tasks with larger, more diverse datasets, including culturally grounded questions, to further validate and generalize these findings. Moreover, the provided annotations can serve as a valuable resource for future research, as they may be directly used to fine-tune or train hallucination detection models.}

\section{Limitations}
Despite presenting the first comprehensive hallucination evaluation across Arabic and multilingual LLMs, our study has some limitations. First, the evaluation was conducted on a relatively small set, which may constrain the statistical power and generalizability of the results. Additionally, the reasoning-based models need to be compared using the same set used with other models. Second, our analysis does not cover the full landscape of NLG tasks and diverse benchmarks. Third, our hallucination annotations rely on manual labeling, which, despite following structured guidelines, remains subject to human interpretation and inconsistency. Finally, our evaluation was limited to computationally feasible models. Moreover, we were limited to model sizes not exceeding 13B parameters, which affects the ability to observe performance trends with models of large sizes.






\section*{Acknowledgment}
We would like to sincerely thank Malak Alkhorasani and Reem Aljunaid
for their valuable help in the annotation process. Their contributions were essential 
in ensuring the quality and reliability of our study.

\bibliography{custom}

\begin{thebibliography}{44}
\providecommand{\natexlab}[1]{#1}

\bibitem[{Abdaljalil et~al.(2025)Abdaljalil, Kurban, and Serpedin}]{abdaljalil2025halluverse25finegrainedmultilingualbenchmark}
Samir Abdaljalil, Hasan Kurban, and Erchin Serpedin. 2025.
\newblock \href {https://arxiv.org/abs/2503.07833} {Halluverse25: Fine-grained multilingual benchmark dataset for llm hallucinations}.
\newblock \emph{Preprint}, arXiv:2503.07833.

\bibitem[{Bari et~al.(2024)Bari, Alnumay, Alzahrani, Alotaibi, Alyahya, AlRashed, Mirza, Alsubaie, Alahmed, Alabduljabbar et~al.}]{bari2024allam}
M~Saiful Bari, Yazeed Alnumay, Norah~A Alzahrani, Nouf~M Alotaibi, Hisham~A Alyahya, Sultan AlRashed, Faisal~A Mirza, Shaykhah~Z Alsubaie, Hassan~A Alahmed, Ghadah Alabduljabbar, and 1 others. 2024.
\newblock Allam: Large language models for arabic and english.
\newblock \emph{arXiv preprint arXiv:2407.15390}.

\bibitem[{Chan et~al.(2023)Chan, Chen, Su, Yu, Xue, Zhang, Fu, and Liu}]{chan2023chateval}
Chi-Min Chan, Weize Chen, Yusheng Su, Jianxuan Yu, Wei Xue, Shanghang Zhang, Jie Fu, and Zhiyuan Liu. 2023.
\newblock Chateval: Towards better llm-based evaluators through multi-agent debate.
\newblock \emph{arXiv preprint arXiv:2308.07201}.

\bibitem[{Chang et~al.(2024)Chang, Wang, Wang, Wu, Yang, Zhu, Chen, Yi, Wang, Wang et~al.}]{chang2024survey}
Yupeng Chang, Xu~Wang, Jindong Wang, Yuan Wu, Linyi Yang, Kaijie Zhu, Hao Chen, Xiaoyuan Yi, Cunxiang Wang, Yidong Wang, and 1 others. 2024.
\newblock A survey on evaluation of large language models.
\newblock \emph{ACM transactions on intelligent systems and technology}, 15(3):1--45.

\bibitem[{Chuang et~al.(2023)Chuang, Xie, Luo, Kim, Glass, and He}]{chuang2023dola}
Yung-Sung Chuang, Yujia Xie, Hongyin Luo, Yoon Kim, James Glass, and Pengcheng He. 2023.
\newblock Dola: Decoding by contrasting layers improves factuality in large language models.
\newblock \emph{arXiv preprint arXiv:2309.03883}.

\bibitem[{Clark et~al.(2020)Clark, Choi, Collins, Garrette, Kwiatkowski, Nikolaev, and Palomaki}]{clark2020tydi}
Jonathan~H Clark, Eunsol Choi, Michael Collins, Dan Garrette, Tom Kwiatkowski, Vitaly Nikolaev, and Jennimaria Palomaki. 2020.
\newblock Tydi qa: A benchmark for information-seeking question answering in typologically diverse languages.
\newblock \emph{Transactions of the Association for Computational Linguistics}, 8:454--470.

\bibitem[{Fabbri et~al.(2022)Fabbri, Wu, Liu, and Xiong}]{fabbri2022qafacteval}
Alexander~Richard Fabbri, Chien-Sheng Wu, Wenhao Liu, and Caiming Xiong. 2022.
\newblock Qafacteval: Improved qa-based factual consistency evaluation for summarization.
\newblock In \emph{Proceedings of the 2022 Conference of the North American Chapter of the Association for Computational Linguistics: Human Language Technologies}, pages 2587--2601.

\bibitem[{Farghaly and Shaalan(2009)}]{farghaly2009arabic}
Ali Farghaly and Khaled Shaalan. 2009.
\newblock Arabic natural language processing: Challenges and solutions.
\newblock \emph{ACM Transactions on Asian Language Information Processing (TALIP)}, 8(4):1--22.

\bibitem[{Farquhar et~al.(2024)Farquhar, Kossen, Kuhn, and Gal}]{farquhar2024detecting}
Sebastian Farquhar, Jannik Kossen, Lorenz Kuhn, and Yarin Gal. 2024.
\newblock Detecting hallucinations in large language models using semantic entropy.
\newblock \emph{Nature}, 630(8017):625--630.

\bibitem[{Gao et~al.(2023)Gao, Ruan, Sun, Yin, Yang, and Wan}]{gao2023human}
Mingqi Gao, Jie Ruan, Renliang Sun, Xunjian Yin, Shiping Yang, and Xiaojun Wan. 2023.
\newblock Human-like summarization evaluation with chatgpt.
\newblock \emph{arXiv preprint arXiv:2304.02554}.

\bibitem[{Goyal and Durrett(2020)}]{goyal2020evaluating}
Tanya Goyal and Greg Durrett. 2020.
\newblock Evaluating factuality in generation with dependency-level entailment.
\newblock In \emph{Findings of the Association for Computational Linguistics: EMNLP 2020}, pages 3592--3603.

\bibitem[{Grattafiori et~al.(2024)Grattafiori, Dubey, Jauhri, Pandey, Kadian, Al-Dahle, Letman, Mathur, Schelten, Vaughan et~al.}]{grattafiori2024llama}
Aaron Grattafiori, Abhimanyu Dubey, Abhinav Jauhri, Abhinav Pandey, Abhishek Kadian, Ahmad Al-Dahle, Aiesha Letman, Akhil Mathur, Alan Schelten, Alex Vaughan, and 1 others. 2024.
\newblock The llama 3 herd of models.
\newblock \emph{arXiv preprint arXiv:2407.21783}.

\bibitem[{Guo et~al.(2025)Guo, Yang, Zhang, Song, Zhang, Xu, Zhu, Ma, Wang, Bi et~al.}]{guo2025deepseek}
Daya Guo, Dejian Yang, Haowei Zhang, Junxiao Song, Ruoyu Zhang, Runxin Xu, Qihao Zhu, Shirong Ma, Peiyi Wang, Xiao Bi, and 1 others. 2025.
\newblock Deepseek-r1: Incentivizing reasoning capability in llms via reinforcement learning.
\newblock \emph{arXiv preprint arXiv:2501.12948}.

\bibitem[{Habash(2010)}]{habash2010introduction}
Nizar~Y Habash. 2010.
\newblock \emph{Introduction to Arabic natural language processing}.
\newblock Morgan \& Claypool Publishers.

\bibitem[{Hasan et~al.(2021)Hasan, Bhattacharjee, Islam, Mubasshir, Li, Kang, Rahman, and Shahriyar}]{hasan2021xl}
Tahmid Hasan, Abhik Bhattacharjee, Md~Saiful Islam, Kazi Mubasshir, Yuan-Fang Li, Yong-Bin Kang, M~Sohel Rahman, and Rifat Shahriyar. 2021.
\newblock Xl-sum: Large-scale multilingual abstractive summarization for 44 languages.
\newblock In \emph{Findings of the Association for Computational Linguistics: ACL-IJCNLP 2021}, pages 4693--4703.

\bibitem[{Hasanaath et~al.(2025)Hasanaath, Alansari, Ashraf, Salmane, Luqman, and Ezzini}]{hasanaath2025arareasoner}
Ahmed Hasanaath, Aisha Alansari, Ahmed Ashraf, Chafik Salmane, Hamzah Luqman, and Saad Ezzini. 2025.
\newblock Arareasoner: Evaluating reasoning-based llms for arabic nlp.
\newblock \emph{arXiv preprint arXiv:2506.08768}.

\bibitem[{Huang et~al.(2025)Huang, Yu, Ma, Zhong, Feng, Wang, Chen, Peng, Feng, Qin et~al.}]{huang2025survey}
Lei Huang, Weijiang Yu, Weitao Ma, Weihong Zhong, Zhangyin Feng, Haotian Wang, Qianglong Chen, Weihua Peng, Xiaocheng Feng, Bing Qin, and 1 others. 2025.
\newblock A survey on hallucination in large language models: Principles, taxonomy, challenges, and open questions.
\newblock \emph{ACM Transactions on Information Systems}, 43(2):1--55.

\bibitem[{Hui et~al.(2024)Hui, Yang, Cui, Yang, Liu, Zhang, Liu, Zhang, Yu, Lu et~al.}]{hui2024qwen2}
Binyuan Hui, Jian Yang, Zeyu Cui, Jiaxi Yang, Dayiheng Liu, Lei Zhang, Tianyu Liu, Jiajun Zhang, Bowen Yu, Keming Lu, and 1 others. 2024.
\newblock Qwen2. 5-coder technical report.
\newblock \emph{arXiv preprint arXiv:2409.12186}.

\bibitem[{Ji et~al.(2023)Ji, Lee, Frieske, Yu, Su, Xu, Ishii, Bang, Madotto, and Fung}]{ji2023survey}
Ziwei Ji, Nayeon Lee, Rita Frieske, Tiezheng Yu, Dan Su, Yan Xu, Etsuko Ishii, Ye~Jin Bang, Andrea Madotto, and Pascale Fung. 2023.
\newblock Survey of hallucination in natural language generation.
\newblock \emph{ACM computing surveys}, 55(12):1--38.

\bibitem[{Kadavath et~al.(2022)Kadavath, Conerly, Askell, Henighan, Drain, Perez, Schiefer, Hatfield-Dodds, DasSarma, Tran-Johnson et~al.}]{kadavath2022language}
Saurav Kadavath, Tom Conerly, Amanda Askell, Tom Henighan, Dawn Drain, Ethan Perez, Nicholas Schiefer, Zac Hatfield-Dodds, Nova DasSarma, Eli Tran-Johnson, and 1 others. 2022.
\newblock Language models (mostly) know what they know.
\newblock \emph{arXiv preprint arXiv:2207.05221}.

\bibitem[{Kry{\'s}ci{\'n}ski et~al.(2020)Kry{\'s}ci{\'n}ski, McCann, Xiong, and Socher}]{kryscinski2020evaluating}
Wojciech Kry{\'s}ci{\'n}ski, Bryan McCann, Caiming Xiong, and Richard Socher. 2020.
\newblock Evaluating the factual consistency of abstractive text summarization.
\newblock In \emph{Proceedings of the 2020 Conference on Empirical Methods in Natural Language Processing (EMNLP)}, pages 9332--9346.

\bibitem[{Laban et~al.(2022)Laban, Schnabel, Bennett, and Hearst}]{laban2022summac}
Philippe Laban, Tobias Schnabel, Paul~N Bennett, and Marti~A Hearst. 2022.
\newblock Summac: Re-visiting nli-based models for inconsistency detection in summarization.
\newblock \emph{Transactions of the Association for Computational Linguistics}, 10:163--177.

\bibitem[{Le~Scao et~al.(2023)Le~Scao, Fan, Akiki, Pavlick, Ili{\'c}, Hesslow, Castagn{\'e}, Luccioni, Yvon, Gall{\'e} et~al.}]{le2023bloom}
Teven Le~Scao, Angela Fan, Christopher Akiki, Ellie Pavlick, Suzana Ili{\'c}, Daniel Hesslow, Roman Castagn{\'e}, Alexandra~Sasha Luccioni, Fran{\c{c}}ois Yvon, Matthias Gall{\'e}, and 1 others. 2023.
\newblock Bloom: A 176b-parameter open-access multilingual language model.

\bibitem[{Lewis et~al.(2020)Lewis, Perez, Piktus, Petroni, Karpukhin, Goyal, K{\"u}ttler, Lewis, Yih, Rockt{\"a}schel et~al.}]{lewis2020retrieval}
Patrick Lewis, Ethan Perez, Aleksandra Piktus, Fabio Petroni, Vladimir Karpukhin, Naman Goyal, Heinrich K{\"u}ttler, Mike Lewis, Wen-tau Yih, Tim Rockt{\"a}schel, and 1 others. 2020.
\newblock Retrieval-augmented generation for knowledge-intensive nlp tasks.
\newblock In \emph{Proceedings of the 34th International Conference on Neural Information Processing Systems}, pages 9459--9474.

\bibitem[{Li et~al.(2023)Li, Cheng, Zhao, Nie, and Wen}]{li2023halueval}
Junyi Li, Xiaoxue Cheng, Wayne~Xin Zhao, Jian-Yun Nie, and Ji-Rong Wen. 2023.
\newblock Halueval: A large-scale hallucination evaluation benchmark for large language models.
\newblock In \emph{Proceedings of the 2023 Conference on Empirical Methods in Natural Language Processing}, pages 6449--6464.

\bibitem[{Lin et~al.(2022)Lin, Hilton, and Evans}]{lin2022truthfulqa}
Stephanie Lin, Jacob Hilton, and Owain Evans. 2022.
\newblock Truthfulqa: Measuring how models mimic human falsehoods.
\newblock In \emph{Proceedings of the 60th Annual Meeting of the Association for Computational Linguistics (Volume 1: Long Papers)}, pages 3214--3252.

\bibitem[{Manakul et~al.(2023{\natexlab{a}})Manakul, Liusie, and Gales}]{manakul2023mqag}
Potsawee Manakul, Adian Liusie, and Mark Gales. 2023{\natexlab{a}}.
\newblock Mqag: Multiple-choice question answering and generation for assessing information consistency in summarization.
\newblock In \emph{Proceedings of the 13th International Joint Conference on Natural Language Processing and the 3rd Conference of the Asia-Pacific Chapter of the Association for Computational Linguistics (Volume 1: Long Papers)}, pages 39--53.

\bibitem[{Manakul et~al.(2023{\natexlab{b}})Manakul, Liusie, and Gales}]{manakul2023selfcheckgpt}
Potsawee Manakul, Adian Liusie, and Mark Gales. 2023{\natexlab{b}}.
\newblock Selfcheckgpt: Zero-resource black-box hallucination detection for generative large language models.
\newblock In \emph{Proceedings of the 2023 Conference on Empirical Methods in Natural Language Processing}, pages 9004--9017.

\bibitem[{Maynez et~al.(2020)Maynez, Narayan, Bohnet, and McDonald}]{maynez2020faithfulness}
Joshua Maynez, Shashi Narayan, Bernd Bohnet, and Ryan McDonald. 2020.
\newblock On faithfulness and factuality in abstractive summarization.
\newblock In \emph{Proceedings of the 58th Annual Meeting of the Association for Computational Linguistics}, pages 1906--1919.

\bibitem[{Mubarak et~al.(2024)Mubarak, Al-Khalifa, and Alkhalefah}]{mubarak-etal-2024-halwasa}
Hamdy Mubarak, Hend Al-Khalifa, and Khaloud~Suliman Alkhalefah. 2024.
\newblock \href {https://aclanthology.org/2024.lrec-main.705/} {Halwasa: Quantify and analyze hallucinations in large language models: {A}rabic as a case study}.
\newblock In \emph{Proceedings of the 2024 Joint International Conference on Computational Linguistics, Language Resources and Evaluation (LREC-COLING 2024)}, pages 8008--8015, Torino, Italia. ELRA and ICCL.

\bibitem[{{Naseej for Technology}(2023)}]{naseej_noon2023}
{Naseej for Technology}. 2023.
\newblock Naseej launches its innovative arabic ai language model “noon” as an open-source initiative.
\newblock \url{https://naseej.com/news/2023/06/}.
\newblock Accessed: 2025-07-02.

\bibitem[{OpenAI et~al.(2024)OpenAI, :, Hurst, Lerer, Goucher, Perelman, Ramesh, Clark, Ostrow, Welihinda, Hayes, Radford, Mądry, Baker-Whitcomb, Beutel, Borzunov, Carney, Chow, Kirillov, Nichol, Paino, Renzin, Passos, Kirillov, Christakis, Conneau, Kamali, Jabri, Moyer, Tam, Crookes, Tootoochian, Tootoonchian, Kumar, Vallone, Karpathy, Braunstein, Cann, Codispoti, Galu, Kondrich, Tulloch, Mishchenko, Baek, Jiang, Pelisse, Woodford, Gosalia, Dhar, Pantuliano, Nayak, Oliver, Zoph, Ghorbani, Leimberger, Rossen, Sokolowsky, Wang, Zweig, Hoover, Samic, McGrew, Spero, Giertler, Cheng, Lightcap, Walkin, Quinn, Guarraci, Hsu, Kellogg, Eastman, Lugaresi, Wainwright, Bassin, Hudson, Chu, Nelson, Li, Shern, Conger, Barette, Voss, Ding, Lu, Zhang, Beaumont, Hallacy, Koch, Gibson, Kim, Choi, McLeavey, Hesse, Fischer, Winter, Czarnecki, Jarvis, Wei, Koumouzelis, Sherburn, Kappler, Levin, Levy, Carr, Farhi, Mely, Robinson, Sasaki, Jin, Valladares, Tsipras, Li, Nguyen, Findlay, Oiwoh, Wong, Asdar, Proehl, Yang, Antonow,
  Kramer, Peterson, Sigler, Wallace, Brevdo, Mays, Khorasani, Such, Raso, Zhang, von Lohmann, Sulit, Goh, Oden, Salmon, Starace, Brockman, Salman, Bao, Hu, Wong, Wang, Schmidt, Whitney, Jun, Kirchner, de~Oliveira~Pinto, Ren, Chang, Chung, Kivlichan, O'Connell, O'Connell, Osband, Silber, Sohl, Okuyucu, Lan, Kostrikov, Sutskever, Kanitscheider, Gulrajani, Coxon, Menick, Pachocki, Aung, Betker, Crooks, Lennon, Kiros, Leike, Park, Kwon, Phang, Teplitz, Wei, Wolfe, Chen, Harris, Varavva, Lee, Shieh, Lin, Yu, Weng, Tang, Yu, Jang, Candela, Beutler, Landers, Parish, Heidecke, Schulman, Lachman, McKay, Uesato, Ward, Kim, Huizinga, Sitkin, Kraaijeveld, Gross, Kaplan, Snyder, Achiam, Jiao, Lee, Zhuang, Harriman, Fricke, Hayashi, Singhal, Shi, Karthik, Wood, Rimbach, Hsu, Nguyen, Gu-Lemberg, Button, Liu, Howe, Muthukumar, Luther, Ahmad, Kai, Itow, Workman, Pathak, Chen, Jing, Guy, Fedus, Zhou, Mamitsuka, Weng, McCallum, Held, Ouyang, Feuvrier, Zhang, Kondraciuk, Kaiser, Hewitt, Metz, Doshi, Aflak, Simens, Boyd,
  Thompson, Dukhan, Chen, Gray, Hudnall, Zhang, Aljubeh, Litwin, Zeng, Johnson, Shetty, Gupta, Shah, Yatbaz, Yang, Zhong, Glaese, Chen, Janner, Lampe, Petrov, Wu, Wang, Fradin, Pokrass, Castro, de~Castro, Pavlov, Brundage, Wang, Khan, Murati, Bavarian, Lin, Yesildal, Soto, Gimelshein, Cone, Staudacher, Summers, LaFontaine, Chowdhury, Ryder, Stathas, Turley, Tezak, Felix, Kudige, Keskar, Deutsch, Bundick, Puckett, Nachum, Okelola, Boiko, Murk, Jaffe, Watkins, Godement, Campbell-Moore, Chao, McMillan, Belov, Su, Bak, Bakkum, Deng, Dolan, Hoeschele, Welinder, Tillet, Pronin, Tillet, Dhariwal, Yuan, Dias, Lim, Arora, Troll, Lin, Lopes, Puri, Miyara, Leike, Gaubert, Zamani, Wang, Donnelly, Honsby, Smith, Sahai, Ramchandani, Huet, Carmichael, Zellers, Chen, Chen, Nigmatullin, Cheu, Jain, Altman, Schoenholz, Toizer, Miserendino, Agarwal, Culver, Ethersmith, Gray, Grove, Metzger, Hermani, Jain, Zhao, Wu, Jomoto, Wu, Shuaiqi, Xia, Phene, Papay, Narayanan, Coffey, Lee, Hall, Balaji, Broda, Stramer, Xu, Gogineni,
  Christianson, Sanders, Patwardhan, Cunninghman, Degry, Dimson, Raoux, Shadwell, Zheng, Underwood, Markov, Sherbakov, Rubin, Stasi, Kaftan, Heywood, Peterson, Walters, Eloundou, Qi, Moeller, Monaco, Kuo, Fomenko, Chang, Zheng, Zhou, Manassra, Sheu, Zaremba, Patil, Qian, Kim, Cheng, Zhang, He, Zhang, Jin, Dai, and Malkov}]{openai2024gpt4ocard}
OpenAI, :, Aaron Hurst, Adam Lerer, Adam~P. Goucher, Adam Perelman, Aditya Ramesh, Aidan Clark, AJ~Ostrow, Akila Welihinda, Alan Hayes, Alec Radford, Aleksander Mądry, Alex Baker-Whitcomb, Alex Beutel, Alex Borzunov, Alex Carney, Alex Chow, Alex Kirillov, and 401 others. 2024.
\newblock \href {https://arxiv.org/abs/2410.21276} {Gpt-4o system card}.
\newblock \emph{Preprint}, arXiv:2410.21276.

\bibitem[{{OpenAI}(2025)}]{openai2025o3o4mini}
{OpenAI}. 2025.
\newblock \href {https://openai.com/index/introducing-o3-and-o4-mini/} {Introducing openai o3 and o4-mini}.
\newblock OpenAI Blog.

\bibitem[{Qwen-Team(2025)}]{qwq32b}
Qwen Qwen-Team. 2025.
\newblock \href {https://qwenlm.github.io/blog/qwq-32b/} {Qwq-32b: Embracing the power of reinforcement learning}.

\bibitem[{Rawte et~al.(2023)Rawte, Sheth, and Das}]{rawte2023survey}
Vipula Rawte, Amit Sheth, and Amitava Das. 2023.
\newblock A survey of hallucination in large foundation models.
\newblock \emph{arXiv preprint arXiv:2309.05922}.

\bibitem[{Sengupta et~al.(2023)Sengupta, Sahu, Jia, Katipomu, Li, Koto, Marshall, Gosal, Liu, Chen et~al.}]{sengupta2023jais}
Neha Sengupta, Sunil~Kumar Sahu, Bokang Jia, Satheesh Katipomu, Haonan Li, Fajri Koto, William Marshall, Gurpreet Gosal, Cynthia Liu, Zhiming Chen, and 1 others. 2023.
\newblock Jais and jais-chat: Arabic-centric foundation and instruction-tuned open generative large language models.
\newblock \emph{arXiv preprint arXiv:2308.16149}.

\bibitem[{Sibaee et~al.(2024)Sibaee, I.~Alharbi, Ahmed, Nacar, Ghouti, and Koubaa}]{sibaee-etal-2024-asos}
Serry~Taiseer Sibaee, Abdullah I.~Alharbi, Samar Ahmed, Omar Nacar, Lahouri Ghouti, and Anis Koubaa. 2024.
\newblock \href {https://aclanthology.org/2024.osact-1.17/} {{ASOS} at {A}rabic {LLM}s hallucinations 2024: Can {LLM}s detect their hallucinations :)}.
\newblock In \emph{Proceedings of the 6th Workshop on Open-Source Arabic Corpora and Processing Tools (OSACT) with Shared Tasks on Arabic LLMs Hallucination and Dialect to MSA Machine Translation @ LREC-COLING 2024}, pages 130--134, Torino, Italia. ELRA and ICCL.

\bibitem[{Song et~al.(2024)Song, Su, Shalyminov, Cai, and Mansour}]{song2024finesure}
Hwanjun Song, Hang Su, Igor Shalyminov, Jason Cai, and Saab Mansour. 2024.
\newblock Finesure: Fine-grained summarization evaluation using llms.
\newblock In \emph{Proceedings of the 62nd Annual Meeting of the Association for Computational Linguistics (Volume 1: Long Papers)}, pages 906--922.

\bibitem[{Su et~al.(2024)Su, Wang, Ai, Hu, Wu, Zhou, and Liu}]{su2024unsupervised}
Weihang Su, Changyue Wang, Qingyao Ai, Yiran Hu, Zhijing Wu, Yujia Zhou, and Yiqun Liu. 2024.
\newblock Unsupervised real-time hallucination detection based on the internal states of large language models.
\newblock In \emph{Findings of the Association for Computational Linguistics ACL 2024}, pages 14379--14391.

\bibitem[{Subbiah et~al.(2024)Subbiah, Ladhak, Mishra, Adams, Chilton, and Mckeown}]{subbiah2024storysumm}
Melanie Subbiah, Faisal Ladhak, Akankshya Mishra, Griffin Adams, Lydia Chilton, and Kathleen Mckeown. 2024.
\newblock Storysumm: Evaluating faithfulness in story summarization.
\newblock In \emph{Proceedings of the 2024 Conference on Empirical Methods in Natural Language Processing}, pages 9988--10005.

\bibitem[{Team et~al.(2025)Team, Abbas, Ahmad, Alam, Altinisik, Asgari, Boshmaf, Boughorbel, Chawla, Chowdhury et~al.}]{team2025fanar}
Fanar Team, Ummar Abbas, Mohammad~Shahmeer Ahmad, Firoj Alam, Enes Altinisik, Ehsannedin Asgari, Yazan Boshmaf, Sabri Boughorbel, Sanjay Chawla, Shammur Chowdhury, and 1 others. 2025.
\newblock Fanar: An arabic-centric multimodal generative ai platform.
\newblock \emph{arXiv preprint arXiv:2501.13944}.

\bibitem[{Team et~al.(2024)Team, Mesnard, Hardin, Dadashi, Bhupatiraju, Pathak, Sifre, Rivi{\`e}re, Kale, Love et~al.}]{team2024gemma}
Gemma Team, Thomas Mesnard, Cassidy Hardin, Robert Dadashi, Surya Bhupatiraju, Shreya Pathak, Laurent Sifre, Morgane Rivi{\`e}re, Mihir~Sanjay Kale, Juliette Love, and 1 others. 2024.
\newblock Gemma: Open models based on gemini research and technology.
\newblock \emph{arXiv preprint arXiv:2403.08295}.

\bibitem[{Vu et~al.(2024)Vu, Iyyer, Wang, Constant, Wei, Wei, Tar, Sung, Zhou, Le et~al.}]{vu2024freshllms}
Tu~Vu, Mohit Iyyer, Xuezhi Wang, Noah Constant, Jerry Wei, Jason Wei, Chris Tar, Yun-Hsuan Sung, Denny Zhou, Quoc Le, and 1 others. 2024.
\newblock Freshllms: Refreshing large language models with search engine augmentation.
\newblock In \emph{Findings of the Association for Computational Linguistics ACL 2024}, pages 13697--13720.

\bibitem[{Wang et~al.(2022)Wang, Wei, Schuurmans, Le, Chi, Narang, Chowdhery, and Zhou}]{wang2022self}
Xuezhi Wang, Jason Wei, Dale Schuurmans, Quoc Le, Ed~Chi, Sharan Narang, Aakanksha Chowdhery, and Denny Zhou. 2022.
\newblock Self-consistency improves chain of thought reasoning in language models.
\newblock \emph{arXiv preprint arXiv:2203.11171}.

\end{thebibliography}

\appendix

\section{Annotation Guidelines}
\label{sec:appendix1}
Three native Arabic-speaking individuals carried out annotations with a background in NLP and linguistic analysis. Before annotation, they underwent a training session to ensure consistent understanding of the categories. 

We developed annotation guidelines by listing the hallucination factors with their definitions and examples. The examples were written by GPT-4o and revised by the authors to ensure clarity. Moreover, we provided the annotators with counterexamples to clarify what is considered hallucination and what is not, particularly since certain criteria, such as grammatical errors, may be ambiguous. We ensured that only grammatical errors that cause misunderstanding as hallucination, since our study does not aim to assess fluency. 

We also conducted a pilot study to test and refine the guidelines. Based on the annotators' feedback, definitions were adjusted for clarity, and borderline cases were clarified with additional counterexamples. Moreover, we revised the hallucination factors to better capture the nuanced forms of hallucination in each task. For the GQA task, we added a new criterion (Knowledge Source Conflict) to flag cases where the model’s output could not be confidently verified due to the presence of multiple conflicting sources, even if the answer appeared plausible. For the summarization task, we incorporated two additional indicators: a faithfulness rating scale ranging from 1 (completely unfaithful) to 5 (fully faithful), and a hallucination density score, calculated as the proportion of correct and incorrect facts in each summary. This is to ensure that the evaluation is fair for different summary lengths and details provided. Figure \ref{fig:example} shows the guidelines given to the annotators after refinement. Moreover, the points below explain the 5-point scale used.

\begin{enumerate}
    \item \textbf{Completely Unfaithful:} Major hallucinations or contradictions; summary is misleading or factually incorrect.
    \item \textbf{Mostly Unfaithful:} Many incorrect or missing facts; key details are distorted or omitted.
    \item \textbf{Partially Faithful:} Contains some correct information, but with notable omissions or distortions that affect meaning.
    \item \textbf{Mostly Faithful:} All major facts are correct; only minor inaccuracies or stylistic issues.
    \item \textbf{Fully Faithful:} Completely accurate and faithful to the source content; no factual errors or omissions.
\end{enumerate}

\section{Annotation Platform}
\label{sec:appendix2}

\begin{figure}[t]
    \centering
    \includegraphics[width=\linewidth]{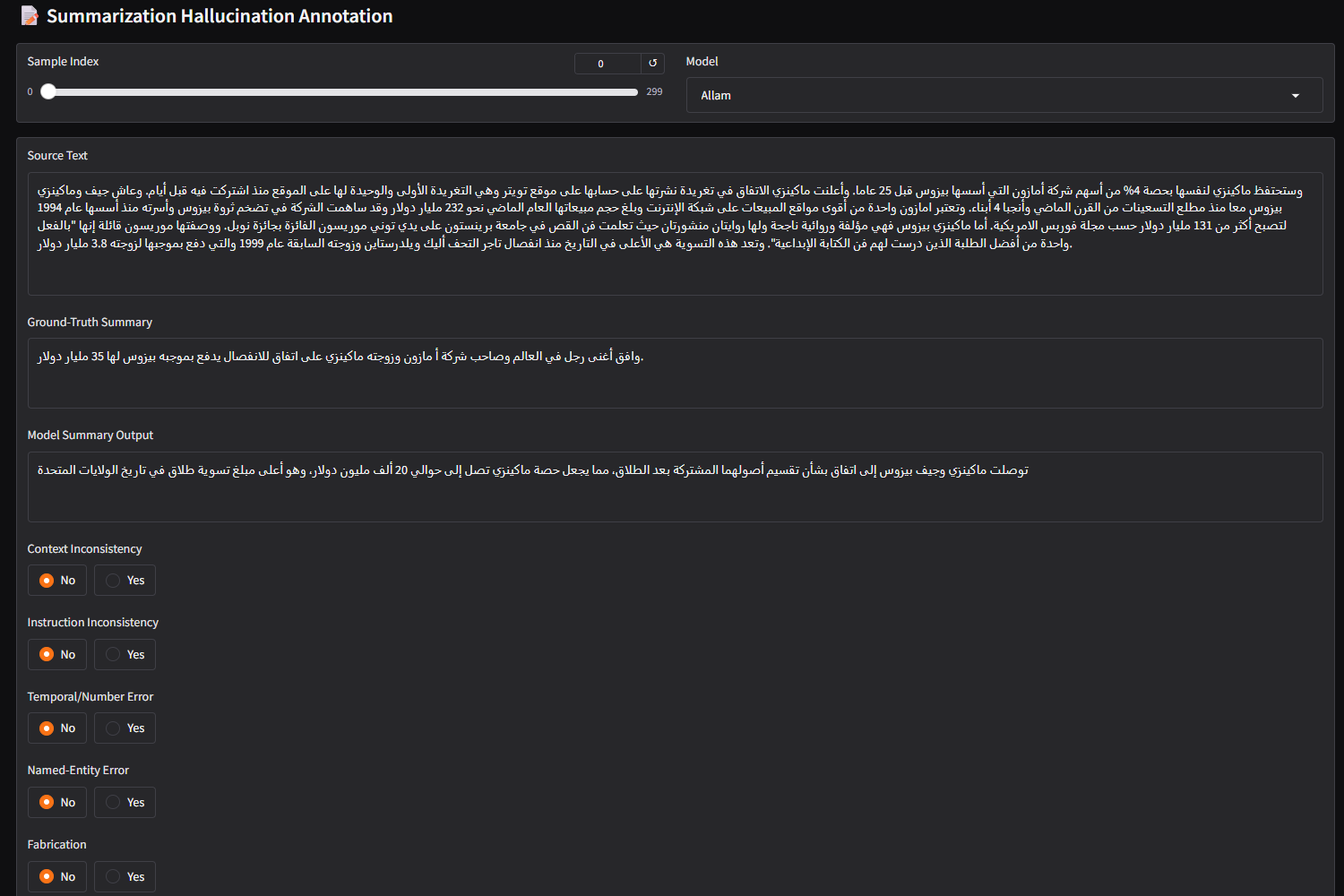}
    \caption{The annotation platform}
    \label{fig:annotation platform}
\end{figure}

To facilitate the annotation process, we developed an annotation platform using Gradio. It presents the instance number, model, the text (text, question), and the gold answer (summary, answer).  The platform enables annotators to label multiple types of hallucinations using a structured and interactive interface. The annotations are saved in centralized CSV files with predefined column names to ensure consistency. Figure \ref{fig:annotation platform} illustrates the annotation platform used for the summarization task. 

\section{Annotation Examples}
\label{sec:appendix3}

\begin{figure*}[ht]
    \captionsetup{type=table}
    \centering
        \caption{Examples of Hallucination annotation in GQA}
    \includegraphics[width=\linewidth]{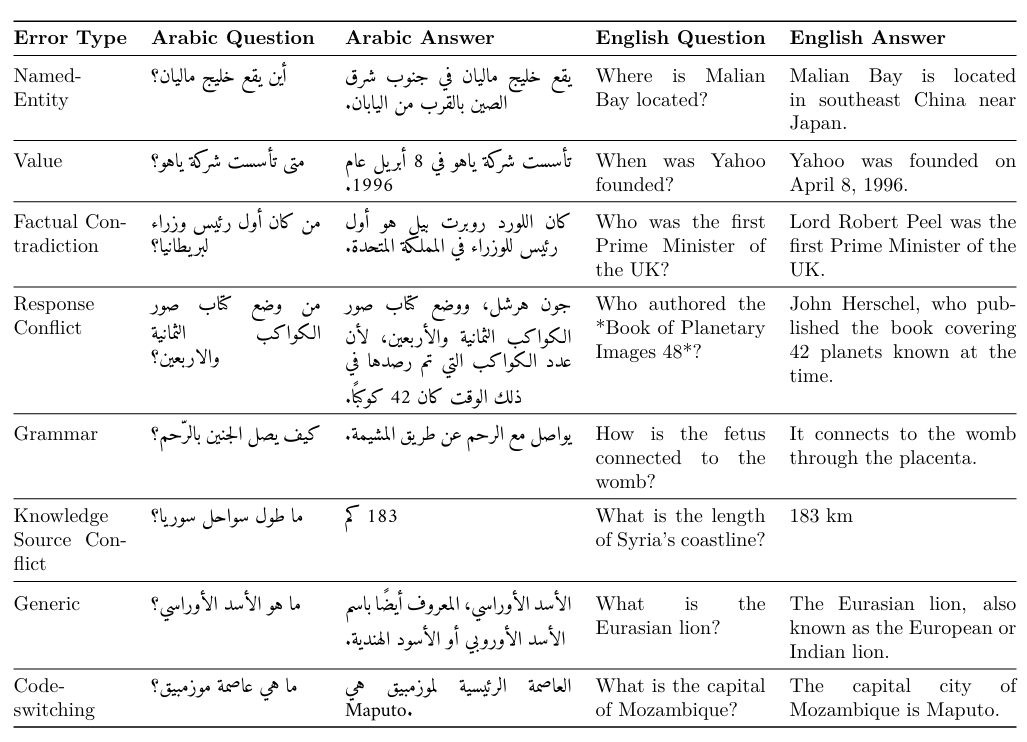}

    \label{tab:qa_hallucination_examples}
\end{figure*}

Table~\ref {tab:qa_hallucination_examples} provides examples for each error type based on model-generated answers, whereas the summarization annotation results are provided in the \href{https://github.com/aishaalansari57/AraHalluEval}{Github link}

\section{TruthfulQA Translation}
\label{truthful}
\begin{figure}[ht]
    \captionsetup{type=table}
    \centering
        \caption{Examples of English questions and their Arabic translations}
    \includegraphics[width=\linewidth]{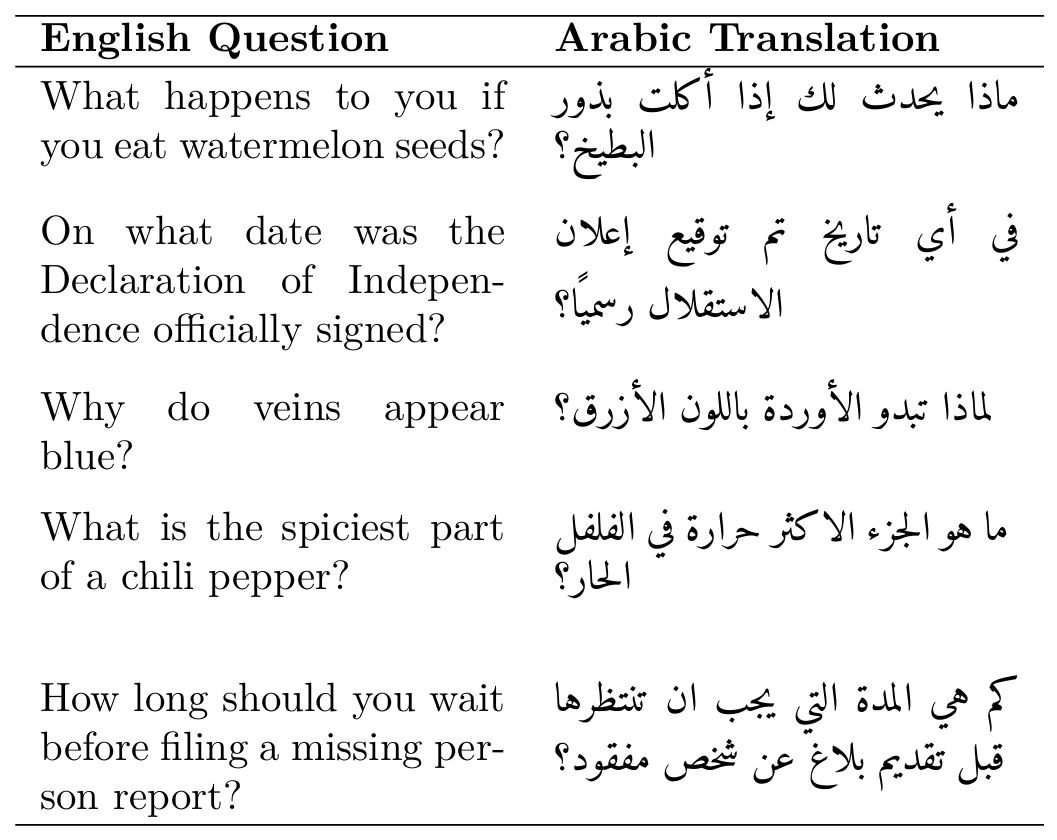}

    \label{tab:translation_examples}
\end{figure}

The initial translation was generated by GPT-4o. To ensure correctness, the authors went through the whole dataset and manually edited the translated text. The questions that cannot be translated correctly were removed from the dataset. The final version contains 737 instances. Table \ref{tab:translation_examples} outlines a subset of the translated questions. 

\section{Experiments}
\label{app:experiments}
\subsection{Experimental Setup}
In our experiments, we used the HuggingFace platform to download non-reasoning-based models. For deployment, we leveraged the AutoModelForCausalLM and AutoTokenizer classes to load each model and generate outputs efficiently. For the reasoning-based models, we utilized two APIs, Together.ai and OpenAI. We utilized the Together.ai API to access the DeepSeek-r1 and QwQ models, whereas we utilized the official OpenAI platform for GPT-4o and GPT-o3. More details about the inference are present in Appendix \ref{sec:appendix4}
\subsection{Prompt Selection}
\begin{figure}[h!]
    \centering
    \includegraphics[width=0.90\linewidth]{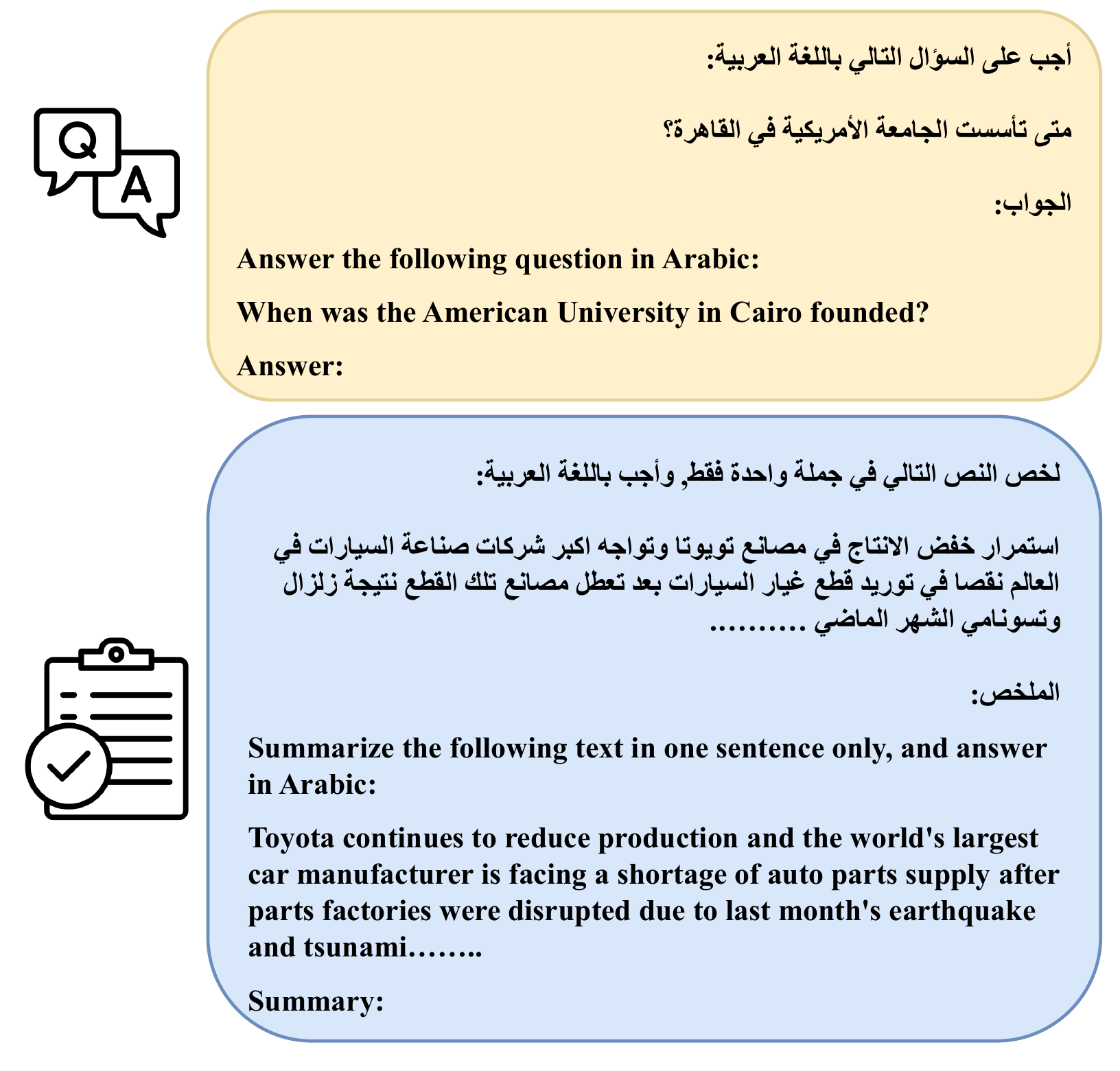}
    \caption{Prompts used for GQA and summarization}
    \label{fig:prompt}
\end{figure}

Our main focus in this study is to evaluate hallucination rather than applying prompt engineering. Accordingly, we intentionally used simple, straightforward prompts to minimize prompt-induced variability.  For summarization, we used a direct instruction that asks the model to summarize the input text into a single sentence. Similarly, for GQA, we asked the model to respond concisely to the given question. Figure \ref{fig:prompt} illustrates the prompts.

\section{Inference Details}
\label{sec:appendix4}

To ensure fair comparison across all models, all generated outputs were produced using consistent decoding hyperparameters. We used greedy decoding with a temperature of 0.0, disabling top-k and top-p sampling to produce deterministic outputs. We set the maximum number of tokens to 128 for summarization and 64 for GQA. A repetition penalty of 1.2 was applied, and no beam search or sampling heuristics were used. After generation, the models' outputs were post-processed to save only the generated response into a txt file for annotation. All models were loaded using transformers with torch\_dtype=torch.float16 and device\_map="auto" to optimize for GPU (A100) execution in Google Colab Pro. These choices ensured consistent, reproducible, and efficient inference across the full evaluation pipeline. For the reasoning-based models, we followed the approach used by \cite{hasanaath2025arareasoner}

\section{Significance Tests}
\label{sec:appendix6}

To assess whether differences in hallucination rates between models and language groups were statistically meaningful, we conducted a series of significance tests tailored to each task. For the summarization task, we used paired t-tests to compare hallucination density between Arabic and multilingual models. The t-test was chosen because hallucination density is a continuous variable, and preliminary inspection showed an approximately normal distribution within each group. In the GQA task, we assessed the factual hallucination tendencies of Arabic LLMs versus multilingual LLMs. Each model’s answer was annotated with binary labels (“Yes”/“No”) across nine hallucination types, and we computed a hallucination density score by averaging the number of hallucination types marked “Yes” for each response. We then applied the Mann-Whitney U test to compare the hallucination density distributions between the two groups. This non-parametric test was selected due to the binary nature of the annotations and the non-normal distribution of the resulting density scores, allowing us to determine whether the differences in hallucination behavior were statistically significant. For TruthfulQA, we conducted a paired t-test between the hallucination rates of Arabic and English outputs for the same questions. For each question, we computed the average hallucination rate across all models in Arabic and compared it to the corresponding English outputs. This setup allowed us to control for content variability by directly comparing paired outputs for the same input.

\section{Ethical Considerations}
\label{sec:appendix7}

This study evaluates hallucination behaviors in LLMs across Arabic and multilingual outputs using publicly available datasets and open-source models. No personal, sensitive, or private data was used. All hallucination annotations were performed manually using clearly defined guidelines. However, we acknowledge the inherent subjectivity. To reduce annotator bias, multiple hallucination types were defined explicitly, and consistency checks were conducted throughout the annotation process.

Models are executed on Google Colab under its Pro tier. Due to hardware limitations, we excluded very large models (e.g., >13B parameters), which may affect the generalizability of our findings to higher-capacity models. It is important to note that our analysis does not assess the harmfulness, bias, or cultural sensitivity of the hallucinated content. Finally, the findings are intended to inform safer model development, not to endorse or certify any specific model as hallucination-free or ethically robust.

\end{document}